\documentclass[runningheads]{llncs}

\usepackage[font=footnotesize,labelfont=bf]{caption}
\usepackage{subcaption}
\usepackage{tikz}
\usepackage{comment}
\usepackage{amsmath,amssymb}
\usepackage{color}

\usepackage[accsupp]{axessibility}

\usepackage{graphicx}
\usepackage{epsfig}

\usepackage{subcaption}
\usepackage{float}
\usepackage[font=small]{caption}
\usepackage{lscape}      
\usepackage{etoolbox} 

\usepackage[ruled, vlined, linesnumbered]{algorithm2e}

\usepackage{microtype}
\usepackage[utf8]{inputenc} 
\usepackage[T1]{fontenc}    
\usepackage{microtype}
\DisableLigatures{encoding = *, family = *}

\usepackage{bm}                          
\usepackage{mathtools}
\usepackage{amsmath, amssymb, amscd}
\usepackage{wasysym} 
\usepackage{nicefrac}       
\usepackage{mdwmath}
\usepackage{eqparbox} 
\usepackage{array} 
\usepackage{nicefrac}

\usepackage{array} 
\usepackage{tabularx}
\usepackage{multirow}
\usepackage{multicol}
\usepackage{booktabs}   

\usepackage{listings}
\usepackage{nameref}
\usepackage{paralist}
\usepackage{enumitem}
\usepackage{xspace}
\usepackage{setspace}

\usepackage[numbers, sort]{natbib}

\usepackage{graphicx}
\usepackage{subcaption}
\usepackage[font=small]{caption}

\usepackage{float}
\usepackage{lscape}

\usepackage{booktabs}
\usepackage{multirow}

\usepackage{paralist}
\usepackage{enumitem}

\usepackage{bm}
\usepackage{epsfig}
\usepackage{mathtools}

\usepackage{color,colortbl}
\usepackage{siunitx} 

\usepackage{comment}

\usepackage{url}  
\usepackage[pagebackref=true,breaklinks=true,colorlinks,urlcolor=blue,citecolor=blue,linkcolor=blue,bookmarks=false]{hyperref}

\usepackage{listings}

\usepackage{xspace}
\usepackage{setspace}

\usepackage{nicefrac}
\usepackage{microtype}
\usepackage[utf8]{inputenc} 
\usepackage[T1]{fontenc}    

\usepackage{url}  
\usepackage{color,colortbl} 

\renewcommand {\vec}[1]{\mathbf{#1}}
\newcommand {\qq}{\vec{q}}
\newcommand {\dqq}{\dot{\vec{q}}}
\newcommand {\ddqq}{\ddot{\vec{q}}}
\newcommand {\duu}{\dot{\vec{u}}}
\newcommand {\dduu}{\ddot{\vec{u}}}
\newcommand {\uu}{\vec{u}}
\newcommand {\HH}{\textbf{H}}
\newcommand {\MM}{\textbf{M}}
\newcommand {\fc}{\vec{f}_\textrm{c}}
\newcommand {\funfc}{f_\textrm{c}}
\newcommand {\fe}{\vec{f}_\textrm{ext}}
\newcommand {\Xh}{\textbf{X}_\textrm{h}}
\newcommand {\xh}{\vec{x}}
\newcommand {\x}{\vec{x}}
\newcommand {\fn}{\vec{f}_n}
\newcommand {\ft}{\vec{f}_t}
\newcommand {\kn}{k_n}
\newcommand {\kf}{k_f}
\newcommand {\vt}{\vec{v}_t}
\newcommand {\xlocal}{\vec{x}_\textrm{local}}
\newcommand {\xclamp}{\vec{x}_\textrm{clamp}}

\newcommand\norm[1]{\Vert#1\Vert}
\newcommand {\qh}{\qq_{\textrm{h}}}
\newcommand {\qobj}{\vec{q}_{\textrm{o}}}

\newcommand {\dqobj}{\dot{\vec{q}}_\textrm{o}}
\newcommand {\fd}{\widehat{\vec{f}}_\textrm{c}}

\newcommand {\Ltask}{\mathcal{L}_\textrm{task}}
\newcommand {\Lphysics}{\mathcal{L}_\textrm{phys}}
\newcommand {\Lgrasp}{\mathcal{L}_\textrm{grasp}}
\newcommand {\uobj}{\uu_\textrm{obj}}
\newcommand {\duobj}{\dot{\vec{u}}_\textrm{o}}
\newcommand {\uh}{\uu_\textrm{h}}
\newcommand {\duh}{\dot{\vec{u}}_\textrm{h}}
\newcommand {\Lqrange}{\mathcal{L}_\textrm{range}}
\newcommand {\Lqlimit}{\mathcal{L}_\textrm{limit}}
\newcommand {\Linter}{\mathcal{L}_\textrm{inter}}
\newcommand {\finter}{\vec{f}_\textrm{link}}

\definecolor{Lightapricot}{rgb}{0.99,0.84,0.69}

\begin{document}
\pagestyle{headings}
\mainmatter
\def\ECCVSubNumber{5242}  

\title{Grasp'D: Differentiable Contact-rich \\ Grasp Synthesis for Multi-fingered Hands} 

\newcommand{\algoName}{Grasp'D\xspace}

\titlerunning{Grasp'D: Differentiable Contact-rich Grasp Synthesis}

\author{
Dylan Turpin\inst{1,2,3} \and
Liquan Wang\inst{1,2,3} \and
Eric Heiden\inst{3} \and
Yun-Chun Chen\inst{1,2} \and \\
Miles Macklin\inst{3} \and
Stavros Tsogkas\inst{4} \and
Sven Dickinson\inst{1,2,4} \and
Animesh Garg\inst{1,2,3}
}

\authorrunning{D. Turpin et al.}

\institute{
$^1\,$University of Toronto, 
$^2\,$Vector Institute, 
$^3\,$Nvidia, 
$^4\,$Samsung  \\
\email{dylanturpin@cs.toronto.edu} 
}

\maketitle

\begin{center}
    \includegraphics[width=\linewidth]{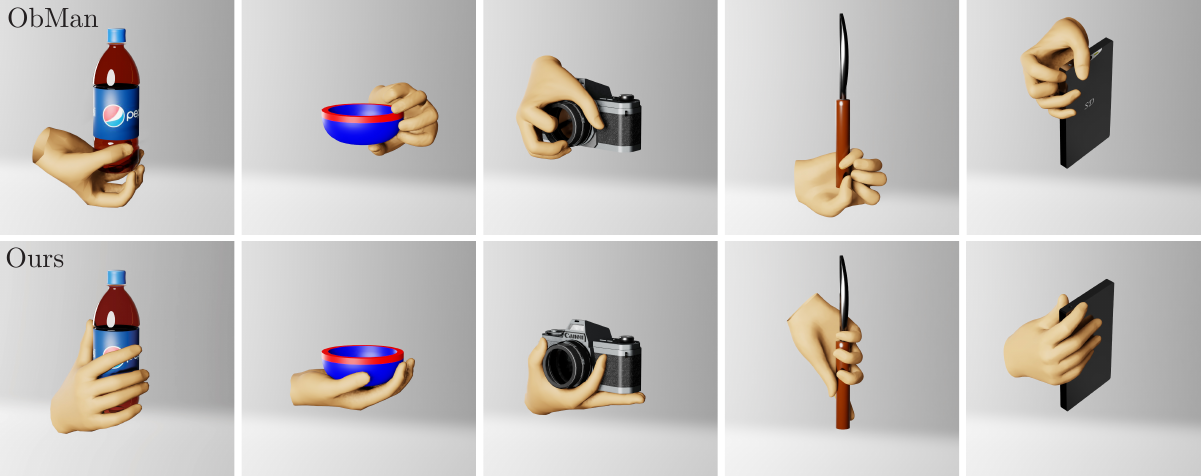}
    \vspace{-5.0mm}
    \captionof{figure}{\small
    \textbf{Multi-finger grasp synthesis with Differentiable Simulation.}
    Analytically synthesized grasps, such as in ObMan~\cite{hasson2019learning} based on the GraspIt!~\cite{miller2004graspit}, plan sparse contacts at the fingertips. 
    Our method (\algoName) for grasp synthesis discovers stable, contact-rich grasps that conform to detailed object surface geometry.
    \algoName creates larger contact-areas that better match the contact distribution of real human grasps.
    }
    \label{fig:1-teaser}
    \vspace{-2.0mm}
\end{center}

\begin{abstract}
The study of hand-object interaction requires generating viable grasp poses for high-dimensional multi-finger models, often relying on analytic grasp synthesis
which tends to produce brittle and unnatural results. 
This paper presents \algoName, an approach for grasp synthesis with a differentiable contact simulation from both known models as well as visual inputs.
We use gradient-based methods as an alternative to sampling-based grasp synthesis, which fails without simplifying assumptions, 
such as pre-specified contact locations and eigengrasps. 
Such assumptions limit grasp discovery and, in particular, exclude high-contact power grasps.
In contrast, our simulation-based approach allows for stable, efficient, physically realistic, high-contact grasp synthesis, even for gripper morphologies with high-degrees of freedom.
We identify and address challenges in making grasp simulation amenable to gradient-based optimization, such as non-smooth object surface geometry, contact sparsity, and a rugged optimization landscape.
\algoName compares favorably to analytic grasp synthesis on human and robotic hand models, 
and resultant grasps achieve over 4× denser contact, leading to significantly higher grasp stability. 
Video and code available at: \href{https://graspd-eccv22.github.io}{graspd-eccv22.github.io}.


\end{abstract}
\section{Introduction}

Humans use their hands to interact with objects of varying shape, size, and material thousands of times throughout a single day.
Despite being effortless -- almost instinctive -- these interactions employ a complex visuomotor system, with components
that correspond to dedicated areas of computer vision research.
Visual inputs from the environment are processed in our brain to recognize objects of interest 
(object recognition~\cite{viola2001rapid, dalal2005histograms, felzenszwalb2009object, girshick2014rich, duan2019centernet}), 
identify modes of interaction to achieve a certain function (affordance prediction~\cite{brahmbhatt2019contactdb,do2018affordancenet,lau2016tactile,porzi2016learning,roy2016multi}), 
and position our hand(s) in a way that enables that function (pose estimation~\cite{hamer2009tracking,supanvcivc2018depth,zimmermann2017learning,baek2019pushing,boukhayma20193d,ge20193d}, grasping~\cite{kokic2020learning,fang2018tog-ijrr,turpin2021gift}).
Proficiency in this task comes from accumulated experience in interacting with the same object over time, and readily
extends to new categories or different instances of the same category.

This is an intriguing observation: humans can leverage accumulated knowledge from previous interactions, 
to quickly infer how to successfully manipulate an unknown object, \emph{purely from visual input}.
Granting machines the same ability to directly translate visual cues into plausible grasp predictions can have significant practical implications in the way robotic manipulators interact with novel objects~\cite{saxena2006robotic,fang2018tog-ijrr} or in virtual environments in AR/VR~\cite{de2017human,gammieri2017coupling}.

Grasp prediction has previously been considered in the context of computer vision~\cite{yang2015grasp,nakamura2017complexities,huang2015we,heumer2007grasp} and robotics~\cite{pirk2017understanding}.
It amounts to predicting the base pose (position and rotation) and joint angles of a robotic or human hand
that is stably grasping a given object.
This prediction is usually conditioned on visual inputs, such as RGB(D) images, point clouds, etc., 
and is typically performed online for real-time applications.
Predicting grasps from visual inputs can be naturally posed as a learning problem,
using paired visual data with their respective grasp annotations.
However, capturing and annotating human grasps is laborious and not applicable
to robotic grasping, so researchers often rely on 
datasets of synthetically generated grasps instead (see Table~\ref{tab:compare} for a list of recent works).
Consequently, high-quality datasets of plausible, diverse grasps are crucial for any modern vision system
performing grasp prediction, motivating the development of better methods for grasp synthesis.

Grasp synthesis assumes that the complete object geometry (e.g., mesh) is known, 
and is usually achieved by optimizing over a grasping metric which can be computed analytically or through simulation.
\emph{Analytic metrics} are handcrafted measures of a grasp's quality.
For example, the epsilon metric~\cite{ferrari1992planning} measures the magnitude of the smallest force that can break a grasp, computed as a function of the contact positions and normals that the grasp induces.
While analytic metrics can be computationally faster, they often transfer poorly to the real world.
\emph{Simulation-based metrics}~\cite{eppner2021acronym,kappler2015leveraging,zhou20176dof} measure grasp quality by running a simulation to test grasp effectiveness, e.g., by shaking the object and checking whether it is dropped.
These can achieve a higher degree of physical fidelity, but require more computation.
In both cases, optimization is usually black box, as neither the analytic metric or simulator is differentiable.
Black box optimization can find good grasps in a reasonable number of steps as long as the search space is low-dimensional, 
e.g., when searching the pose space of parallel-jaw grippers~\cite{eppner2021acronym,veres2017integrated,depierre2018jacquard,mousavian20196,eppner2019billion}.
However, when the number of degrees of freedom becomes larger, as in the case of multi-finger grippers, black box optimization over a grasping metric (whether analytic or simulation-based) becomes infeasible.
Simplifying assumptions can be made to reduce the dimensionality of the search space, but they often reduce the plausibility of generated grasps.

\begin{table}[t]
  \centering
  \begin{minipage}[c]{0.49\textwidth}
   \resizebox{\linewidth}{!}
  {
  \begin{tabular}{llll}
    \toprule
    \rowcolor[HTML]{CBCEFB} 
    \textbf{Year} & \textbf{Name} & \textbf{\begin{tabular}[c]{@{}l@{}}Hand \\ Model(s)\end{tabular}} & \textbf{\begin{tabular}[c]{@{}l@{}}Analytic (A) or \\ Human Capture (HC)\end{tabular}} \\ 
    \midrule
    2019 & ObMan~\cite{hasson2019learning}                  & MANO    & A (GraspIt!~\cite{miller2004graspit}) \\
    \rowcolor[HTML]{EFEFEF} 
    2019 & ContactDB~\cite{brahmbhatt2019contactdb}         & MANO    & HC \\
    2020 & Hope-net~\cite{doosti2020hope}                   & MANO    & A (ObMan~\cite{hasson2019learning}) \\
    \rowcolor[HTML]{EFEFEF} 
    2020 & UniGrasp~\cite{shao2020unigrasp}                 & Various & A (FastGrasp~\cite{pokorny2013classical}) \\
    2020 & ContactPose~\cite{brahmbhatt2020contactpose}     & MANO    & HC \\
    \rowcolor[HTML]{EFEFEF} 
    2020 & GANHand~\cite{corona2020ganhand}                 & MANO    & Other (manual) \\
    2020 & Grasping Field~\cite{karunratanakul2020grasping} & MANO    & A (ObMan) \\
    \rowcolor[HTML]{EFEFEF} 
    2020 & GRAB~\cite{taheri2020grab}                       & MANO    & HC \\
    2021 & Multi-Fin GAN~\cite{lundell2020multi}            & Barrett & A (GraspIt!) \\
    \rowcolor[HTML]{EFEFEF} 
    2021 & DDGC~\cite{lundell2021ddgc}                      & Barrett & A (GraspIt!) \\
    2021 & Contact-Consistency~\cite{jiang2021hand}         & MANO    & A (ObMan) \\
    \bottomrule
  \end{tabular}
  }
  \end{minipage}
  \begin{minipage}[c]{0.49\textwidth}
  \vspace{-4.0mm}
  \caption{
  Modern vision-based grasp prediction for multi-finger hands relies on datasets created by human capture or analytic synthesis.
  Human capture is expensive and does not address the need for robotic grasp datasets.
  Analytic synthesis is only practical under significant limiting assumptions 
  that exclude key grasp types~\cite{corona2020ganhand,hasson2019learning}.
  }
  \label{tab:compare}
  \end{minipage}
  \vspace{-6.0mm}
\end{table}

To address these shortcomings, we propose \algoName, a grasp synthesis pipeline based on \emph{differentiable simulation} which can generate contact-rich grasps that realistically conform to object surface geometry without any simplifying assumptions.
A metric based on differentiable simulation admits gradient-based optimization, which is sample-efficient, even in high-dimensional spaces, and affords all the benefits of simulation-based metrics, i.e., physical plausibility, scalability, and extendability.
Differentiable grasping simulation, however, also presents new challenges.
Non-smooth object geometry (e.g., at the edges or corners of a cube) results in discontinuities in the contact forces and, subsequently, our grasping metric, complicating gradient-based optimization.
Adding to that, if the hand and the object are not touching, small perturbations to the hand pose do not generate any additional force, resulting in vanishing gradients.
Finally, the optimization landscape is rugged, making optimization challenging.
Once the hand is touching the object, small changes to the hand pose may result in large changes to contact forces (and our metric).

We address these challenges as follows:
(1) At the start of each optimization, we simulate contact between the hand and a smoothed, padded version of the object surface that gradually resolves to the true, detailed surface geometry, using a coarse-to-fine approach.
This smoothing softens discontinuities in surface normals, allowing gradient-based optimization to smoothly move from one continuous surface area to another. 
This is enabled by our signed-distance function (SDF) approach to collision detection, which lets us freely recover a rounded object surface as the radius $r$ level set of the SDF.
(2) We allow gradients to \textit{leak} through force computations for contact points that are not yet in the collision, introducing a biased gradient that can be followed to create new contacts.
The intuition behind this choice is similar to the one for using LeakyReLU activations to prevent the phenomenon of ``dying neurons'' in deep neural networks~\cite{maas2013rectifier}.
(3) Inspired by Contact-Invariant Optimization (CIO)~\cite{mordatch2012discovery,mordatch2012contact}, we relax the problem formulation by introducing additional force variables that allow physics violations to be treated as a cost rather than a constraint.
In effect, this decomposes the problem into finding contact forces that solve the task (of keeping the object stably in place) and finding a hand pose that provides those forces.
We evaluate our method on synthetic object models from ShapeNet~\cite{chang2015shapenet} and object meshes reconstructed from the YCB RGB-D dataset~\cite{calli2017yale}.
Experimental results show that our method generates contact-rich grasps with physical realism and with favorable performance against an existing analytic method~\cite{hasson2019learning}.

\begin{figure}[t]
  \includegraphics[width=\linewidth]{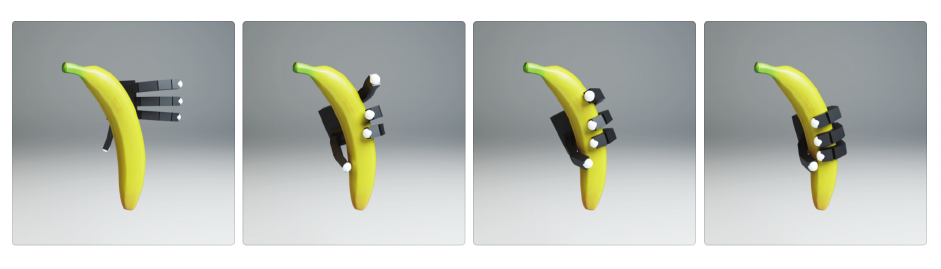}
  \vspace{-8.0mm}
  \captionof{figure}{ 
  Our method can synthesize grasps for both human and robotic hands, such as the four-finger Allegro hand in this figure.
  After  hand initialization, we run gradient-based optimization to iteratively improve the grasp,
  in terms of stability and contact area.
  We include additional examples in Appendix~B.
  }
  \label{fig:allegro}
  \vspace{-5.0mm}
\end{figure}

Figure~\ref{fig:1-teaser} displays example grasps generated by our method side-by-side with grasps from \cite{hasson2019learning}.
Because we do not make assumptions about contact locations
or reduce the dimensionality of the search space,
our method can discover contact-rich grasps that are more stable and more plausible than the fingertip-only grasps usually discovered by analytic synthesis.
The same procedure works equally for robotic hands.
Figure~\ref{fig:allegro} displays snapshots of an optimization trajectory for an Allegro hand.
As optimization progresses and our simulated metric decreases, the grasp becomes increasingly stable, plausible, and high-contact.

\vspace{-4.0mm}
\subsubsection{Summary of contributions:}
\begin{enumerate}[topsep=0pt, leftmargin=*]
    \item We propose a differentiable simulation-based protocol for generating synthetic grasps from visual data. 
    Unlike other simulation-based approaches, our method can scale to tens of thousands of dense contacts, and 
    discover plausible, contact-rich grasps, without any simplifying assumptions.
    \item We address challenges arising from the differentiable nature of our scheme, using a coarse-to-fine
    SDF collision detection approach, defining leaky gradients for contact points that are not yet in collision,
    and integrating physics violations as additional terms to our cost function. 
    \item We show that our method finds grasps with better stability, lower interpenetration, and higher contact area when compared to analytic grasp synthesis baselines, and justify our design choices through extensive evaluations. 
\end{enumerate}

\vspace{-1.0mm}
\section{Related Work}
\vspace{-1.0mm}

\noindent \textbf{Grasp synthesis.}
Although analytic metrics have been successfully applied to parallel-jaw gripper grasp synthesis (based on grasp wrench space analysis~\cite{ferrari1992planning,miller2004graspit,goldfeder2009columbia}, robust grasp wrench space analysis~\cite{weisz2012pose,mahler2017dex}, or caging~\cite{rodriguez2012caging,mahler2016energy}), more recent works~\cite{depierre2018jacquard,kappler2015leveraging,mousavian20196,eppner2021acronym} have focused on simulation-based synthesis.
While they are more computationally costly, simulation-based metrics for parallel-jaw grasps better align with human judgement~\cite{kappler2015leveraging} and with real world performance~\cite{mahler2019learning,danielczuk2019reach,mousavian20196,eppner2021acronym}.
In contrast to parallel-jaw grippers, multi-finger grasp synthesis is still largely analytic, with many recent works in multi-finger robotic grasping~\cite{shao2020unigrasp,lundell2020multi,lundell2021ddgc}, grasp affordance prediction~\cite{karunratanakul2020grasping}, and hand-object pose estimation~\cite{hasson2019learning,doosti2020hope,jiang2021hand} relying on datasets of analytically synthesized grasps (see Table~\ref{tab:compare}).
Notably, \cite{lundell2020multi,lundell2021ddgc,karunratanakul2020grasping,hasson2019learning,doosti2020hope,jiang2021hand} all use datasets synthesized with the GraspIt!~\cite{miller2004graspit} simulator, which is widely used for both multi-finger robotic and human grasp synthesis.
The ObMan dataset~\cite{hasson2019learning} for hand-object pose estimation (also used in \cite{karunratanakul2020grasping,jiang2021hand}) is constructed by performing grasp synthesis with the MANO hand~\cite{MANO:SIGGRAPHASIA:2017} in the GraspIt! Eigengrasp planner, 
and rendering the synthesized grasps against realistic backgrounds.
The GraspIt! Eigengrasp planner optimizes analytic metrics based on grasp wrench space analysis.
Dimensionality reduction~\cite{ciocarlie2007dexterous} in the hand joint space, 
or using pre-specified contact locations for each hand link
can be used to make the problem more tractable, but this limits the space of discoverable grasps
and requires careful tuning.
Our approach can successfully operate in the full grasp space, eschewing such simplifying assumptions
while excelling in terms of physical fidelity over analytic synthesis for multi-finger grippers.

\vspace{2.0mm}
\noindent \textbf{Human grasp capture.}
To estimate human grasps from visual inputs, existing methods train models on large-scale datasets~\cite{brahmbhatt2019contactdb,brahmbhatt2020contactpose,hampali2020honnotate,garcia2018first,tzionas2016capturing}.
Collecting these datasets puts humans in a lab with precise, calibrated cameras, lidar, and special gloves for accurately capturing human grasp poses.
A human in the loop may also be needed for collecting annotations. 
All these requirements make the data collection process expensive and laborious.
In addition, the captured grasps are only appropriate for human hands and not for robotic ones 
(which are important for many applications~\cite{chen2022system,allshire2021transferring}).
Some works~\cite{lakshmipathy2022contact,brahmbhatt2019contactgrasp} aim to transfer human grasps to robotic hands by matching contact patterns, but these suffer from important limitations, since the same contacts
may not be achievable by human and robotic hands, given
differences in their morphology and articulation constraints
(e.g., see Fig. 8 of~\cite{lakshmipathy2022contact}).
Our method provides a procedural way of generating high quality grasps for any type of hand -- human or robotic.

\vspace{2.0mm}
\noindent \textbf{Vision-based grasp prediction.}
Whereas grasp synthesis is useful for generating grasps when full object geometry is available (i.e., a mesh or complete SDF is given), practical scenarios require predicting grasps from visual input.
GANHand~\cite{corona2020ganhand} learns to predict human grasp affordances (as poses of a MANO~\cite{MANO:SIGGRAPHASIA:2017} hand model) from input RGBD images using GANs.
Since analytic synthesized datasets do not include many high-contact grasps,
the authors also released the YCB Affordance dataset of 367 fine-grained grasps of the YCB object set~\cite{calli2017yale}, created by manually setting MANO hand joint angles in the GraspIt! simulator's GUI.
Rather than predicting joint angles, Grasping Field~\cite{karunratanakul2020grasping} takes an implicit approach to grasp representation by learning to jointly predict signed distances for the MANO hand and the object to be grasped.
For parallel-jaw grippers, most recent works~\cite{mahler2019learning,mousavian20196,sundermeyer2021contact,jiang2021synergies} learn from simulation-based datasets (e.g., \cite{kappler2015leveraging,eppner2021acronym}).
In contrast, multi-finger grasp prediction systems are still trained on either analytically synthesized datasets or datasets of captured human grasps (see Table~\ref{tab:compare}).
\cite{lundell2020multi,lundell2021ddgc,karunratanakul2020grasping,hasson2019learning,doosti2020hope,jiang2021hand} all use analytically synthesized datasets from the GraspIt! simulator~\cite{miller2004graspit}, whereas
\cite{brahmbhatt2019contactdb,brahmbhatt2020contactpose,taheri2020grab} use datasets of captured human grasps.
\cite{grady2021contactopt,jiang2021hand} use captured human grasps to train a contact model, then refine grasps at test-time by optimizing hand pose  to match predicted contacts.
The higher quality training data generated by our grasp synthesis pipeline can lead to improved performance for 
any of these vision-based grasping prediction systems.
Our system can also be used directly for vision-based grasp prediction, 
by running simulations with reconstructed objects (see Section~\ref{exp:E2}).

\vspace{2.0mm}
\noindent \textbf{Differentiable Grasping.}
We know of two works that have created differentiable grasp metrics in order to take advantage of 
gradient-based optimization for multi-finger grasp synthesis.
\cite{liu2020deep} formulates a differentiable version of the epsilon metric~\cite{ferrari1992planning} and uses it to synthesize grasps with the shadow robotic hand.
They formulate the epsilon metric computation as a semidefinite programming (SDP) problem.
Sensitivity analysis on this problem can then provide the gradient of the solution with respect to the problem parameters, including gripper pose.  They manually label $45$ potential contact points on the gripper.
In contrast, we are able to scale to tens of thousands of contact points.
Since the gripper may not yet be in contact with the object, they use an exponential weighting of points.
Liu et al.~\cite{liu2021synthesizing} formulate a differentiable force closure metric and use gradient-based optimization to synthesize grasps with the MANO~\cite{MANO:SIGGRAPHASIA:2017} hand model.
Their formulation assumes zero friction and that the magnitude
of all contact forces is uniform across contact points (although an error term
allows both of these constraints to be slightly violated).
Our method requires neither of these assumptions:
the user can specify varying friction coefficients, and
contact forces at different points are free to vary realistically.
Their optimization problem involves finding a hand pose and a 
subset of candidate contact points on the hand that minimize an energy function.
They find that the algorithm performs better with a smaller number
of contact points and candidates.
Selecting 3 contact points from the 773 candidate vertices of the MANO hand, it takes about 40 minutes to find 5 acceptable grasps.
In contrast, our method is able to scale to tens of thousands of contact points while synthesizing an acceptable grasp in about 5 minutes.
Notably, both of these prior works aim to take an analytic metric (the epsilon metric~\cite{ferrari1992planning}) and make a differentiable variant.
In contrast, we are presenting a differentiable simulation-based metric, which prior work on parallel-jaw grippers suggests will have greater
physical fidelity~\cite{danielczuk2019reach,mousavian20196,eppner2021acronym} and better match human judgements~\cite{kappler2015leveraging} than analytic metrics.

\vspace{2.0mm}
\noindent \textbf{Differentiable Physics.}
There has been significant progress in the development of differentiable 
physics engines~\cite{hu2019chainqueen, hu2019difftaichi, geilinger2020add, brax2021github, werling2021fast, qiao2021efficient, heiden2021neuralsim, heiden2021disect,xie2022shac}.
However, certain limitations in recent approaches render them inadequate.
Brax~\cite{brax2021github} and the Tiny Differentiable Simulator~\cite{heiden2021neuralsim} only support collision primitives and cannot model general collisions between objects.
Nimblephysics~\cite{werling2021fast} supports mesh-to-mesh collision, but cannot handle cases where the gradient of
contact normals with respect to position is zero (e.g., on a mesh face).
While its analytic computation of gradients is fast, Nimblephysics requires manually 
writing forward and backward passes in C\texttt{++}, and only runs on CPU.
Our work presents a new class of differentiable physics simulators to addresses many of these shortcomings. 
Further, \algoName supports GPU parallelism, enabling us to scale to tens of thousands of contacts, effectively approximating surface contacts. 
\begin{figure}[!t]
    \centering
    \includegraphics[width=\linewidth]{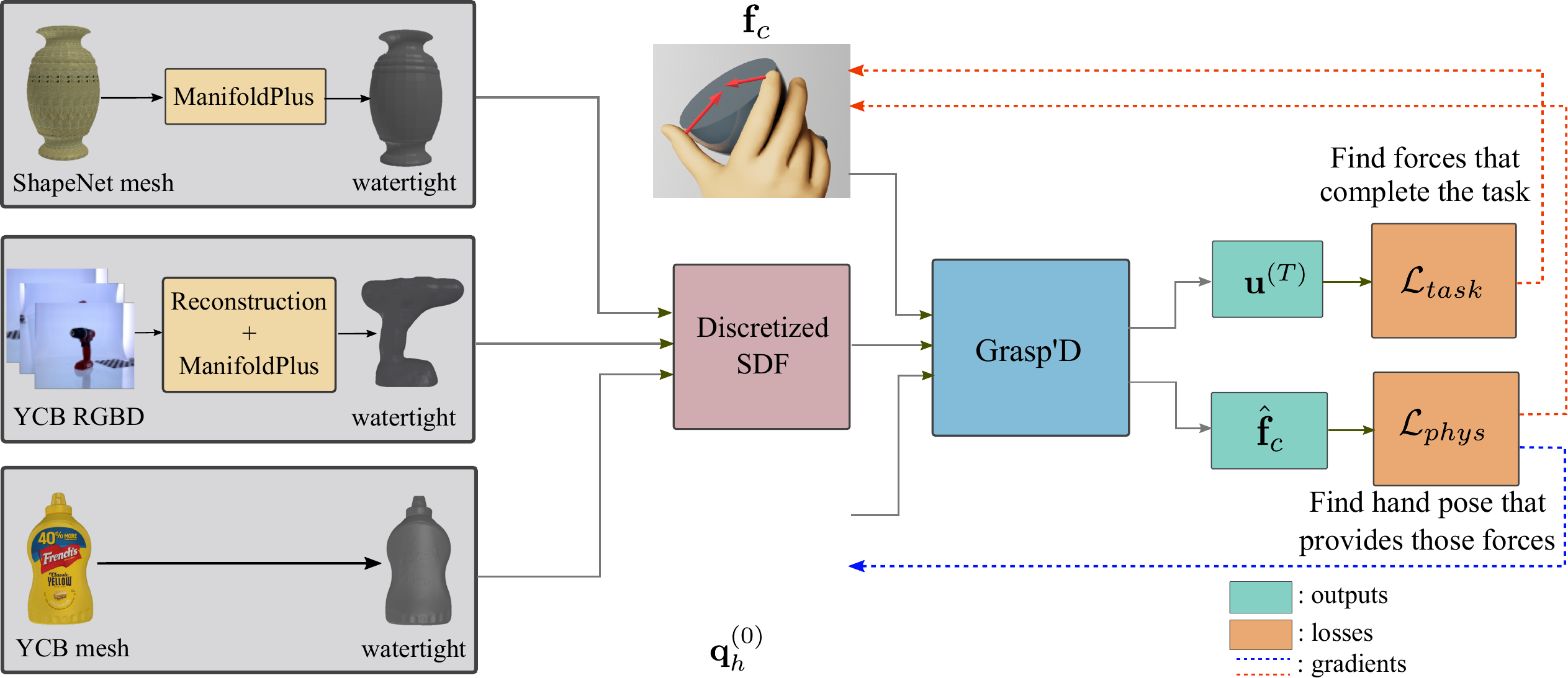}
    \caption{
    \textbf{Method overview.}
\algoName takes as input the discretized-SDF of an object 
(computed from a mesh or reconstructed from RGB-D) and synthesizes a stable grasp
that can hold the object static as we vary the object's initial velocity.
We optimize jointly over a hand pose $\mathbf{u}^{(T)}$ and the stabilizing forces $\hat{\mathbf{f}}_{c}$ provided by its contacts.
    }
    \label{fig:2-method}
    \vspace{-5.0mm}
\end{figure}

\vspace{-2.0mm}
\section{\algoName: Differentiable Contact-rich Grasp Synthesis } \label{sec:method}
\vspace{-2.0mm}

We present a method for solving the grasp synthesis problem (Figure~\ref{fig:2-method}).
From an input object and hand model (represented respectively by a signed-distance function and an articulation chain with mesh links),
we generate a physically-plausible stable grasp, as a base pose and joint angles of the hand.
This is achieved by iterative gradient-based optimization over a metric computed by differentiable simulation.
The final grasp is dependent on the pose initialization of the hand,
so different grasps can be recovered by sampling different starting poses.
We detail our method below, but first outline the challenges that motivate our design.

\vspace{2.0mm}
\noindent \textbf{Non-smooth object geometry.}
When optimizing the location of contacts between a hand and a sphere, the gradient of contact normals with respect to contact positions is well-defined and continuous, allowing gradient-based optimization to smoothly adjust contact positions along the sphere surface.
But most objects are not perfectly smooth.
Discontinuities in surface normals (e.g., at the edges or corners of a cube) result in discontinuities in contact normals and their gradients with respect to contact positions.
Gradient-based optimization cannot effectively optimize across these discontinuities (e.g., cannot follow the gradient to move contact locations from one face of a cube to another).
We address this with a coarse-to-fine smoothing approach, optimizing against a smoothed and padded version of the object surface that gradually resolves to the true surface as optimization continues (see Section \ref{subsec:obj_model}).

\vspace{2.0mm}
\noindent \textbf{Contact sparsity.}
Of all possible contacts between the hand and object, only a sparse subset is active at any given time.
If a particular point on the hand is inactive (not in contact with the object), then an infinitesimal perturbation of the hand pose will not change its status 
(make it touch the object).
The gradient of the force applied by any inactive contact (with respect to hand pose) will be exactly zero.
This means that gradient-based optimization can not effectively create new contacts, since contacts that are not already active do not contribute to the gradient.
We address this by allowing gradient to \textit{leak} through the force computations of inactive contacts (see Section \ref{subsec:contact-dynamics}).

\vspace{2.0mm}
\noindent \textbf{Rugged optimization landscape.}
When many contacts are active (i.e., hand touching the object), 
small changes to hand pose may result in large changes to contact forces and, subsequently, large changes to our grasp metric.
This makes gradient-based optimization challenging.
We address this with a problem relaxation inspired by Contact-Invariant Optimization~\cite{mordatch2012discovery,mordatch2012contact} (see Section~\ref{subsec:relaxed-objective}).

\vspace{-2.0mm}
\subsection{Rigid body dynamics}

In the interest of speed and simplicity, we limit ourselves to simple rigid body dynamics.
Let $\qq$ and $\uu$ be the joint and spatial coordinates, respectively, with first and second time derivatives $\dqq$, $\ddqq$, $\duu$, $\dduu$.
Let $\MM$ be the mass matrix.
The kinematic map $\HH$ maps joint coordinate time derivatives to spatial velocities as $\dqq = \HH(\qq)\uu$, and is related to contact and external forces ($\fc$ and $\fe$) through the following motion equation: $\HH\MM\HH^\top\ddqq = \fc + \fe$,
which yields the semi-implicit Euler update used for discrete time stepping~\cite{bender2014interactive}:
\begin{align}
\dqq^{(t+1)} &\leftarrow  \dqq^{(t)} + \Delta t \MM^{-1} (\fc + \fe)\label{eq:update1}\\
\qq^{(t+1)} &\leftarrow \qq^{(t)} + \Delta t \dqq^{(t+1)}\label{eq:update2}.
\end{align}

\vspace{-2.0mm}
\subsection{Object model with coarse-to-fine surface smoothing}
\label{subsec:obj_model}

\noindent \textbf{SDF representation.}
For the purpose of collision detection, the hand is represented by a set of surface points $\Xh$, and
the object to grasp is represented by its Signed Distance Function (SDF), $\phi(\x)$ (similar to~\cite{fuhrmann2003distance,macklin2020local,bender2014continuous}).
The SDF maps a spatial position $\x \in \mathbb{R}^3$ to its distance to the closest point on the surface of the object,
with a negative or positive sign for interior and exterior points, respectively~\cite{osher2006level}.
The object surface can be recovered as the zero level-set of the SDF: $\{\x | \phi(\x) = 0\}$.
The gradient of the SDF $\nabla \phi(\x)$ is always of unit magnitude,
corresponds to the surface normal for $\xh$ on the object surface,
and yields the closest point on the object as $\xh - \phi(\xh)\nabla\phi(\xh)$.
SDF representations are well-suited to differentiable collision detection~\cite{macklin2020local},
since contact forces can be written in terms of a penetration depth ($\phi$)
and normal direction ($\nabla \phi$),
for which gradients can be computed as $\nabla \phi$ and $\nabla^2 \phi$, respectively.

Whereas primitive objects (e.g., a sphere or box) admit an analytic SDF,
this is not the case for complex objects, for which an SDF representation is not readily available.
We model the object to be grasped by a discretized SDF which we extract from ground truth meshes (easier to come by for most object sets~\cite{calli2017yale,chang2015shapenet}), yielding a 3D grid.
Given a query point $\x$, to compute $\phi(\x)$ based on the grid, we first convert $\x$ to local shape coordinates
(where the object is in canonical pose: unrotated and centered at the origin), yielding $\xlocal$.
If $\xlocal$ falls within the bounds of the grid, we map it to grid indices and compute $\phi(\xlocal)$ by tri-linear interpolation of neighbouring grid cells.
If $\xlocal$ falls outside the grid, we clamp it to the grid bounds, yielding $\xclamp$,
and compute $\phi(\x) := \phi(\xclamp) + \norm{\x - \xclamp}$.

\vspace{2.0mm}
\noindent \textbf{Coarse-to-fine smoothing.}
To successfully optimize contact locations over non-smooth object geometry
we employ surface smoothing in a coarse-to-fine way.
At the start of each optimization, we define the object surface \emph{not} as the zero level-set of the SDF, but as the radius $r$ level-set: $\{\x | \phi(\x) = r > 0\}$,
which gives a smoothed and padded version of the original surface.
As optimization continues, we decrease $r$ on a linear schedule until it reaches $0$, yielding the original surface.
This coarse-to-fine smoothing allows gradient-based optimization to effectively move contact points across discontinuities
and prevents the optimization from quickly overfitting to local geometric features.
We set $r$ to approximately 10cm at the start of each optimization. Details are in Appendix~A.2.

\vspace{-2.0mm}
\subsection{Contact dynamics with leaky gradient}
\label{subsec:contact-dynamics}

\noindent \textbf{Contact forces.}
We use a primal (penalty-based) formulation of contact forces, which allows us to compute derivatives with autodiff~\cite{baydin2018automatic} and keep a consistent memory footprint.
For a given point $\xh \in \Xh$, the resultant contact force is
\begin{align}
    \fc & = \fn + \ft \label{eq:fc}\\
    \fn & = \kn \min(\phi(\xh),\ 0) \nabla \phi(\xh) \label{eq:fn}\\
    \ft & = -\min( \kf \norm{\vt},\ \mu \norm{\fn} ) \vt, \label{eq:ft}
\end{align}
where $\fn$ is the normal component, proportional to penetration depth $\phi(\xh)$,
and $\ft$ is the frictional component, computed using a Coulomb friction model.
$\kn$ and $\kf$ are the normal and frictional stiffness coefficients, respectively,
$\mu$ is the friction coefficient, and $\vt$ is the component of relative velocity between hand and object at the contact point $\xh$ that is tangent to the contact normal $\nabla \phi(\xh)$.

\vspace{2.0mm}
\noindent \textbf{Leaky gradients.}
At any one time, most possible hand-object contacts are inactive -- a property we refer to as \textit{contact sparsity}.
Since an infinitesimal perturbation to hand pose will not activate these contacts (i.e., will not make them touch the object),
the gradient of their contact forces with respect to hand pose is zero, i.e., $\partial\fc / \partial\qq = \partial\fc / \partial\dqq = \partial\fc / \partial\ddqq = 0$.
When the hand is not touching the object, all contacts are inactive and gradient-based optimization can get stuck in a plateau.
We work around this by computing a \textit{leaky} gradient for the normal force term.
From equation \eqref{eq:fn}, we have $\frac{\partial \norm{\fn}}{\partial\qq} = 0$ if $\phi(\x) \geq 0$
but we instead set
\begin{equation}
    \frac{\partial \norm{\fn}}{\partial\qq} := \begin{cases}
        \kn \frac{\partial\phi}{\partial\qq} &\textrm{if $\phi(\x) < 0$}\\
        \alpha \kn \frac{\partial\phi}{\partial\qq} &\textrm{otherwise}
    \end{cases},
\end{equation}
where $\alpha \in [0,1]$ controls how much gradient leaks through the minimum.
We set $\alpha=0.1$ in our experiments. 

\subsection{Grasping metric and problem relaxation}
\label{sec:grasp-metric}

\noindent \textbf{Simulation setup.}
To compute the grasp metric,
we simulate the rigid-body interaction between a hand
and an object.
The hand is kinematic (does 
not react to contact forces), while the object is 
dynamic (thus subject to contact forces).
The simulator state is given by the configuration vector $\qq$ and its first and second time derivatives $\dqq,\ddqq$. 
$\qq$ is composed of hand and object components $\qq = (\qh, \qobj)$ with corresponding spatial coordinates $\uu = (\uh, \uobj)$.
The object is always initialized with the same
configuration $\qobj^{(0)}$: unrotated and untranslated at the origin.
Given a state $\qq^{(t)}$, following equations \eqref{eq:update1} and \eqref{eq:update2}, our simulator uses a semi-implicit Euler update scheme
to compute subsequent state $\qq^{(t+1)}$.

\vspace{2.0mm}
\noindent \textbf{Computing the grasp metric by simulation.}
To measure the quality of a candidate grasp $\qh$,
we test its ability to withstand forces applied to the object.
Given an initial state $\qq^{(0)} = (\qh,\qobj^{(0)})$, we apply an initial velocity ${\dqobj}^{(0)}$ to the object.
The hand is kept static, with  $\duh = 0$.
We run forward simulation to compute the object's final velocity ${\duobj}^{(T)}$.
A stable grasp will produce contact forces that resist
the object velocity, so lower $\norm{{\duobj}^{(T)}}$ indicates a more stable grasp.
In fact, a stable grasp should be able to resist object velocities in \emph{any} direction,
so we perform multiple simulations with different initial velocities
and average the results.
This suggests the following basic grasp metric: 
for each set of $M$ simulations, indexed by $m = \{1, \ldots, M\}$, we set a different initial object velocity, run the simulation, and record $L_m = \norm{{\duobj}^{(T)}}$.
Then, averaging, we have
\begin{equation}
    \Lgrasp = \sum_{m=1}^M \frac{L_m}{M}.
\end{equation}
Since $\Lgrasp$ is a differentiable function of the output of a differentiable simulation,
it is itself differentiable with respect to $\qh$, and we can compute loss
gradients $\partial \Lgrasp / \partial \qh$ and use gradient-based optimization to find stable grasps.

Unfortunately, in practice, this basic procedure does not succeed.
As explained at the beginning of Section~\ref{sec:method},
the grasp optimization landscape is extremely rugged,
with sharp and narrow ridges, peaks, and valleys.
Our leaky contact force gradients (see Section~\ref{subsec:contact-dynamics}) provide some help in escaping plateaus, but once the hand is in contact with the object, small changes in hand configuration still cause large jumps in contact forces by
making/breaking contacts and shifting contact normals.
However, differentiability alone does not resolve this issue.

\vspace{2.0mm}
\noindent \textbf{Problem relaxation.}
\label{subsec:relaxed-objective}
Inspired by Contact-Invariant Optimization~\cite{mordatch2012discovery,mordatch2012contact}
we relax the problem making it more forgiving to gradient-based optimization.
Specifically, we introduce additional \textit{desired} or \textit{prescribed} contact force variables.
This allows us to model physics violations as a cost rather than
a constraint.
For each surface point on the hand $\xh^i \in \Xh$, we introduce a 6-dimensional vector
$\fd^i$ representing the desired hand-object contact wrench arising from contact at $\xh^i$.

Our overall loss now has two components.
The task loss $\Ltask(\fd)$ measures whether the prescribed forces $\fd$ successfully resist initial object velocities.
This is computed identically to the previous $\Lgrasp$, except that
instead of computing contact forces according to equations \eqref{eq:fc}, \eqref{eq:fn} and \eqref{eq:ft},
contact forces are simply set equal to $\fd$.
The physics violation loss $\Lphysics(\qh,\fd)$ measures whether the hand configuration $\qh$ actually provides the desired forces $\fd$.
It is computed as
\begin{equation}
    \Lphysics(\qh, \fd) = \norm{\funfc(\qh) - \fd},
\end{equation}
where $\funfc(\qh)$ is the contact force arising from the hand pose $\qh$ according to equations \eqref{eq:fc}, \eqref{eq:fn} and \eqref{eq:ft}.

Intuitively, minimizing these losses corresponds to finding a set of desired forces (as close as possible to the
actual contact forces arising from the current hand configuration) that complete the task,
and finding a hand configuration that provides those forces.
We expect problem formulations derived from and inspired by Contact-Invariant Optimization~\cite{ciocarlie2007dexterous,ciocarlie2009hand}
to be a fruitful area of research as they are made newly attractive by advances in 
differentiable simulation.

\vspace{2.0mm}
\noindent \textbf{Additional heuristic losses.}
We include some additional losses that improve the plausibility of resulting grasps.
Most hand models have defined joint range limits.
Let $\qh^{\textrm{low}}$ and $\qh^{\textrm{up}}$ be the lower and upper joint limits respectively.
$\Lqrange$ encourages hand joints to be near the middle of their ranges.
$\Lqlimit$ penalizes hand joints outside of their range.
$\Linter$ penalizes self intersections of the hand.
\begin{align}
    \Lqrange(\qh) &= \norm{\qh - \frac{ \qh^{\textrm{up}} + \qh^{\textrm{low}}}{2}}\\
    \Lqlimit(\qh) &= \max(\qh - \qh^{\textrm{up}}, 0) + \max(\qh^{\textrm{low}} - \qh, 0)\\
    \Linter(\qh) &= \norm{\finter}.
\end{align}
The hand is kinematic, so it is not subject to contact forces.
However, we still compute forces arising from contact between the hand links,
for use in this loss term, as $\finter$.
We ignore contacts between neighbouring links in the chain.
For the purpose of computing $\finter$,
we represent each hand link as both a point set and an SDF
and compute $\finter$ according to equations \eqref{eq:fc}, \eqref{eq:fn}, and \eqref{eq:ft}.

\vspace{-1.5mm}
\subsection{Optimization}
We use the Modified Differential Multiplier Method~\cite{platt1987constrained}, treating $\Ltask < C_\textrm{task}$ and $\Lqlimit < C_\textrm{limit}$ as constraints,
while minimizing $\Lphysics$, $\Lqlimit$ and $\Linter$.
We update our parameters $\fd$ and $\qh$ using the Adamax~\cite{kingma2014adam} optimizer.
Details of learning rates, $C_\textrm{task}$ and $C_\textrm{limit}$ can be found in Appendix~A.7.
\vspace{-1.0mm}
\section{Experiments}
\vspace{-1.0mm}

Our evaluations and analysis of \algoName answer the following questions:
\begin{enumerate}[leftmargin=*]
  \item How well does \algoName perform compared to analytic methods? (Section~\ref{exp:E1})
  \item Can \algoName generalize to objects reconstructed from real-world RGBD images? (Section~\ref{exp:E2})
  \item How much do coarse-to-fine SDF collision and the problem relaxation contribute to final performance? (Section~\ref{exp:E3})
\end{enumerate}

\subsection{Experimental setup}
For each experiment, we synthesize grasps following the procedure described in Section~\ref{sec:method}.
We compute the metric with $M=3$ simulations:
each setting a different initial velocity on the hand:
$(0,0,0)$, $(0.01,0.01,0.01)$ or $(-0.01,-0.01,-0.01)$m/s.
Each simulation is run for a single timestep of length $1\times10^{-5}$s.

\vspace{2.0mm}
\noindent \textbf{Evaluation metrics.}
We follow~\cite{hasson2019learning} and use contact area (CA), intersection volume (IV), and the ratio between contact area and intersection volume ($\frac{\mathrm{CA}}{\mathrm{IV}}$).
We compute evaluation metrics that measure grasp stability and contact area.
In addition, we measure the contact area each grasp creates and the volume of hand-object interpenetration.
We compute two analytic measures of stability -- the Ferrari-Canny (epsilon $\epsilon$)~\cite{ferrari1992planning} and the volume metric (Vol) -- and one simulated measure: the simulation displacement (SD) metric introduced in~\cite{hasson2019learning}.

\vspace{2.0mm}
\noindent \textbf{Hand parameterization.}
We use a differentiable PyTorch layer~\cite{hasson2019learning} to compute the 773
vertices of the MANO hand~\cite{MANO:SIGGRAPHASIA:2017} model.
The input is a set of weights for principal components extracted from the MANO dataset of human scans~\cite{MANO:SIGGRAPHASIA:2017}.
We find that this PCA parameterization provides a useful prior for human-like hand poses. 
We use the maximum number of principal components (44).

\begin{table}[!t]
  \centering
  \resizebox{0.8\linewidth}{!} 
  {
  \begin{tabular}{l|S[table-format=4.5]S[table-format=4.5]S[table-format=4.5]S[table-format=4.5]S[table-format=4.5]S[table-format=4.5]}
  \toprule
  \rowcolor[HTML]{CBCEFB} 
  \textbf{Method} & \textbf{CA} $\uparrow$ & \textbf{IV} $\downarrow$ & $\frac{\mathrm{\mathbf{CA}}}{\mathrm{\mathbf{IV}}}$ $\uparrow$ & \textbf{$\epsilon$ $\uparrow$} & \textbf{Vol $\uparrow$} & \textbf{SD $\downarrow$} \\ 
  \midrule
  \rowcolor{Lightapricot}
  Scale (Unit) & \footnotesize{$\text{cm}^2$} & \footnotesize{$\text{cm}^3$} & \footnotesize{$\text{cm}^{-1}$} & \footnotesize{$\times10^{-1}$} & \footnotesize{$\times10^{1}$} & \footnotesize{$\text{cm}$} \\ 
  \midrule
  ObMan~\cite{hasson2019learning} (top2) & 9.4 & 1.28 & 7.37 & 4.70 & 1.36 & 1.95  \\
  \rowcolor[HTML]{EFEFEF} 
  ObMan~\cite{hasson2019learning} (top5) & 7.8 & \textbf{1.05} & 7.37 & 4.52 & 1.36 & 2.22 \\
  \algoName (top2) & $\mathbf{43.0}$ & 5.70 & \textbf{7.55} & 5.01 & 1.44 & \textbf{0.59} \\
  \rowcolor[HTML]{EFEFEF} 
  \algoName (top5) & \text{41.4} & 5.48 & \textbf{7.55} & \textbf{5.02} & \textbf{1.46} & 1.04 \\
  \bottomrule
  \end{tabular}
  }
  \small
  \vspace{2.0mm}
  \caption{
  \textbf{Experimental results.}
  We synthesize MANO hand grasps for ShapeNet objects.
  Our grasps achieve over $4\times$ denser contact (as measured by contact surface area - CA) than an analytic synthesis baseline~\cite{hasson2019learning},
  leading to significantly higher grasp stability ($4\times$ lower simulation displacement - SD).
  Higher contact does result in higher interpenetration, but we keep a similar ratio of contact area to interpenetration volume.
  }
  \label{exp:results-main}
  \vspace{-8.0mm}
\end{table}

\vspace{-2.0mm}
\subsection{Grasp synthesis with ShapeNet models} \label{exp:E1}

We compare to baseline grasps from the ObMan~\cite{hasson2019learning} dataset,
which generates grasps with the GraspIt!~\cite{miller2004graspit} simulator using an analytic metric.
We report these metrics over the top-2 and top-5 grasps per scaled object, with ranking decided by simulation displacement for our method and by ObMan's heuristic measure (detailed in Appendix C.2 of~\cite{hasson2019learning}) for theirs.
Further details in Appendix A.6.

\vspace{2.0mm}
\noindent \textbf{Data.}
We evaluate our approach to grasp synthesis by generating grasps with the
MANO human hand~\cite{MANO:SIGGRAPHASIA:2017} model for 57 ShapeNet~\cite{chang2015shapenet} objects that span 8 categories (bottles, bowls, cameras, cans, cellphones, jars, knives, remote controls), and are each considered at 5 different scales (as in ObMan).
See the Appendix~A for details of mesh pre-processing, initialization, simulation, and optimization.

\vspace{2.0mm}
\noindent \textbf{Results.}
Results are presented in Table~\ref{exp:results-main}.
Grasps generated by our method (both top-2 and top-5) have a contact area of around $42\mathrm{cm}^2$. 
This is higher than the $\sim20\mathrm{cm}^2$ area achieved with fingertip only grasps~\cite{brahmbhatt2019contactdb}
and about $4\times$ higher than grasps from the ObMan dataset (top-2 or top-5).
These contact-rich grasps achieve modest improvements in analytic measures of stability, 
and a significant reduction in simulation displacement ($\sim3\times$ for top-2 grasps).
Visualizations of our generated grasps in Figure~\ref{fig:1-teaser} confirm that these grasps achieve larger areas of contact by closely conforming to object surface geometry, whereas the analytically generated grasps largely make use of fingertip contact only.
These higher contact grasps have accordingly higher interpenetration, but the ratio between contact area and intersection volume is similar to the ObMan baseline.

\begin{figure}[t]
  \centering
  \begin{subfigure}[t]{0.15\linewidth}
    \includegraphics[width=\linewidth]{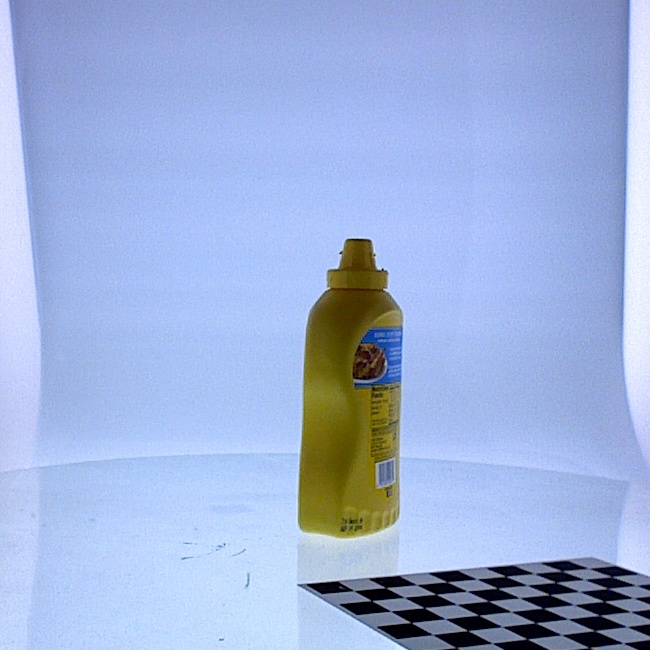}
  \end{subfigure}
  \begin{subfigure}[t]{0.15\linewidth}
    \includegraphics[width=\linewidth]{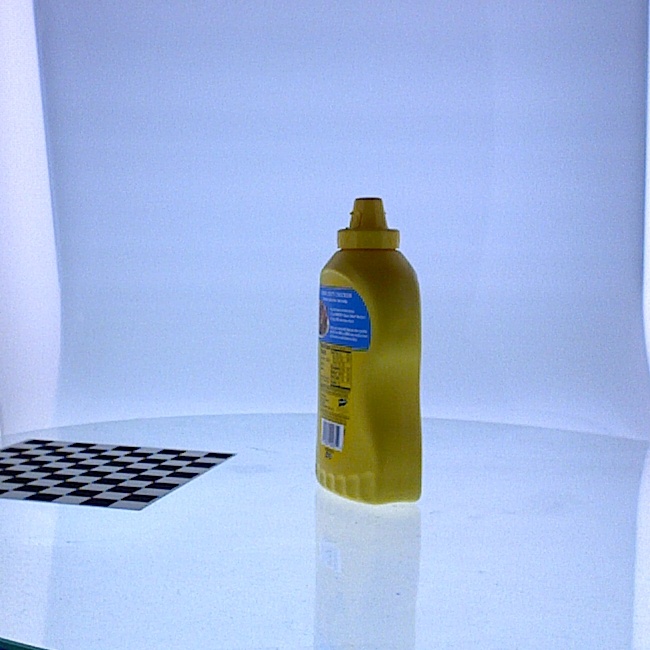}
  \end{subfigure}
  \hfill
  \begin{subfigure}[t]{0.15\linewidth}
    \includegraphics[width=\linewidth]{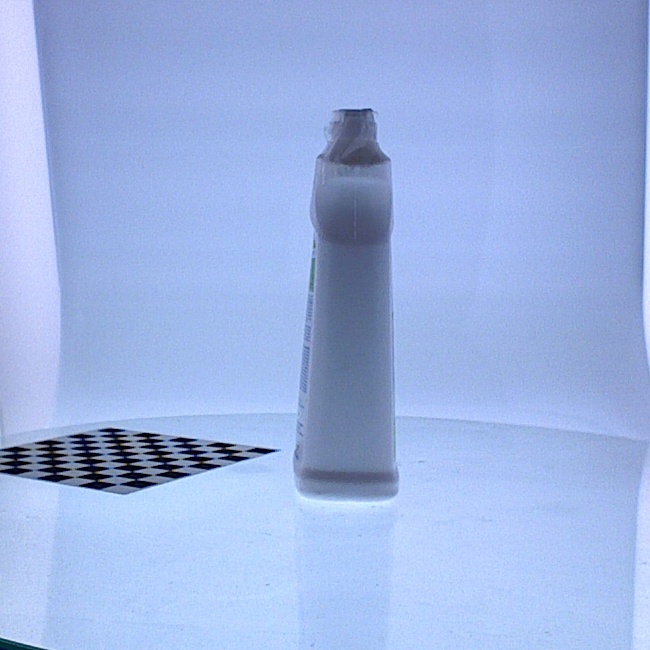}
  \end{subfigure}
  \begin{subfigure}[t]{0.15\linewidth}
    \includegraphics[width=\linewidth]{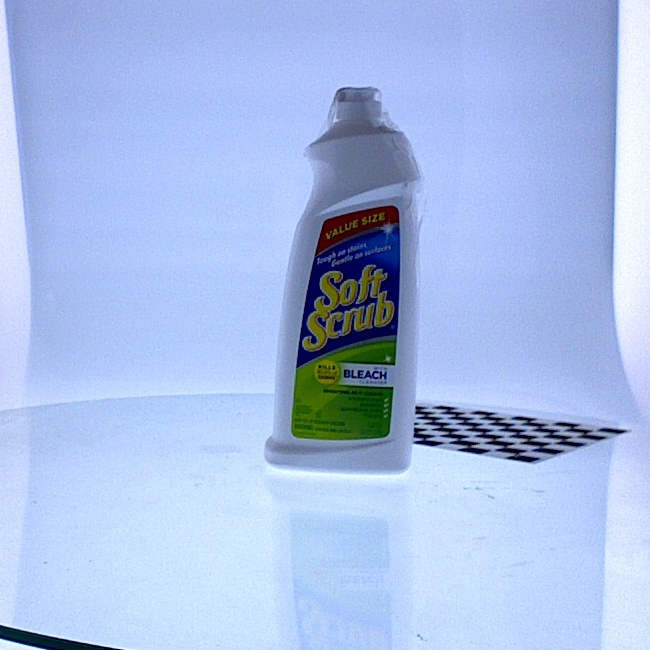}
  \end{subfigure}
  \hfill
  \begin{subfigure}[t]{0.15\linewidth}
    \includegraphics[width=\linewidth]{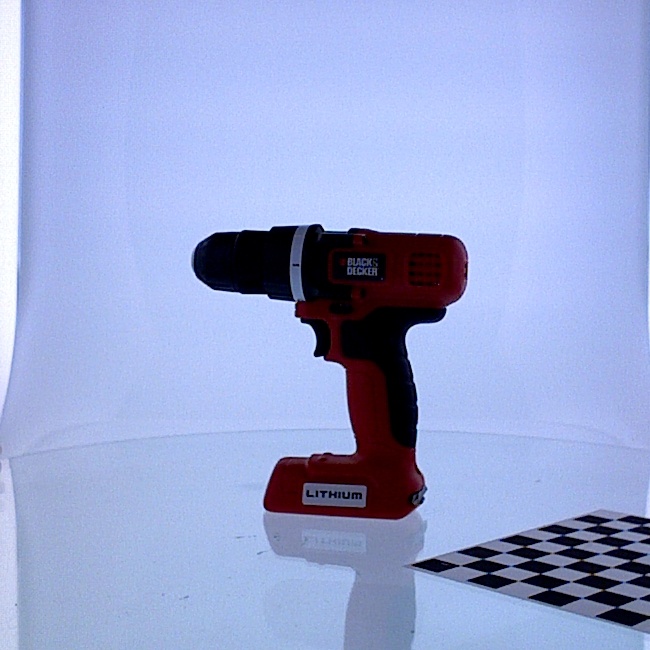}
  \end{subfigure}
  \begin{subfigure}[t]{0.15\linewidth}
    \includegraphics[width=\linewidth]{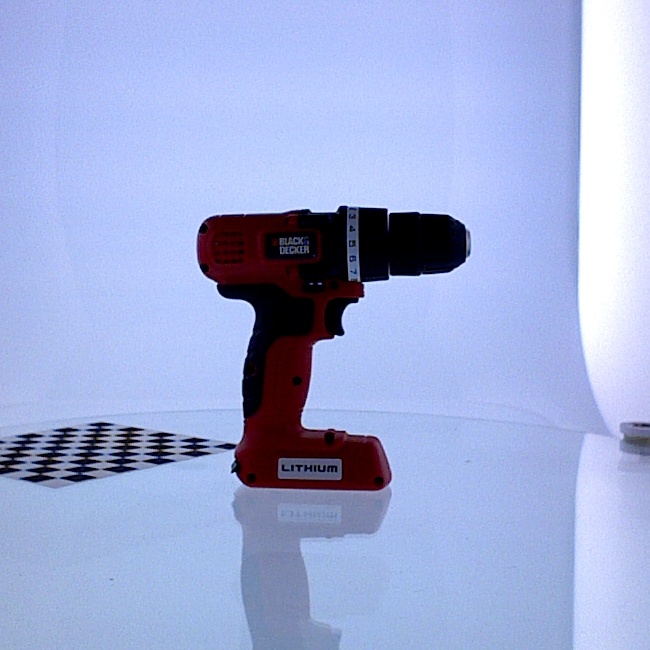}
  \end{subfigure} 
  \\
  \begin{subfigure}[t]{0.155\linewidth}
    \includegraphics[width=\linewidth]{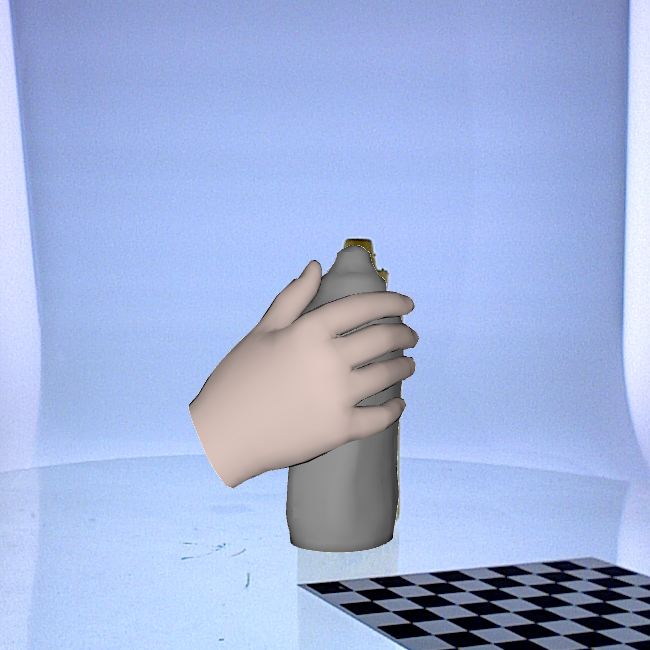}
  \end{subfigure}
  \begin{subfigure}[t]{0.15\linewidth}
    \includegraphics[width=\linewidth]{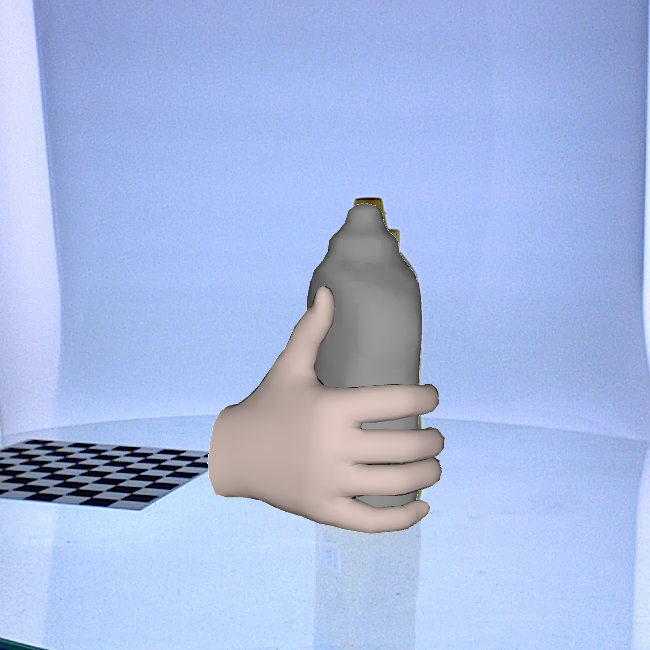}
  \end{subfigure}
  \hfill
  \begin{subfigure}[t]{0.15\linewidth}
    \includegraphics[width=\linewidth]{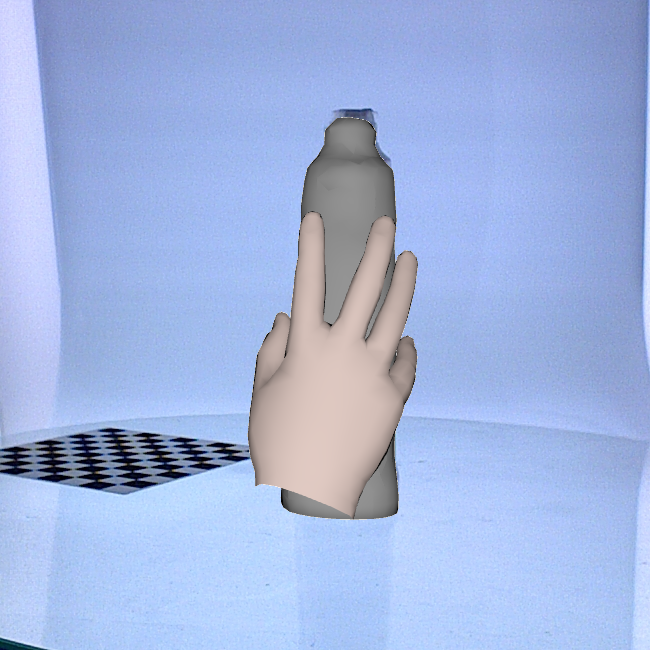}
  \end{subfigure}
  \begin{subfigure}[t]{0.15\linewidth}
    \includegraphics[width=\linewidth]{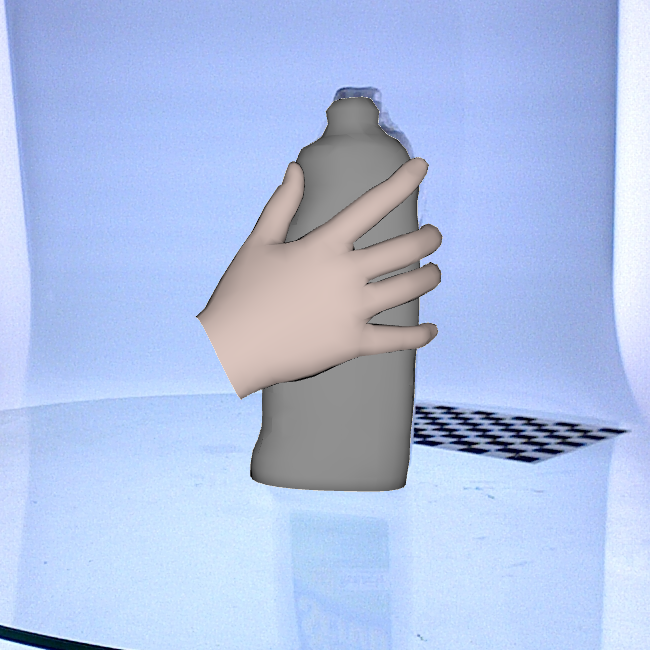}
  \end{subfigure}
  \hfill
  \begin{subfigure}[t]{0.15\linewidth}
    \includegraphics[width=\linewidth]{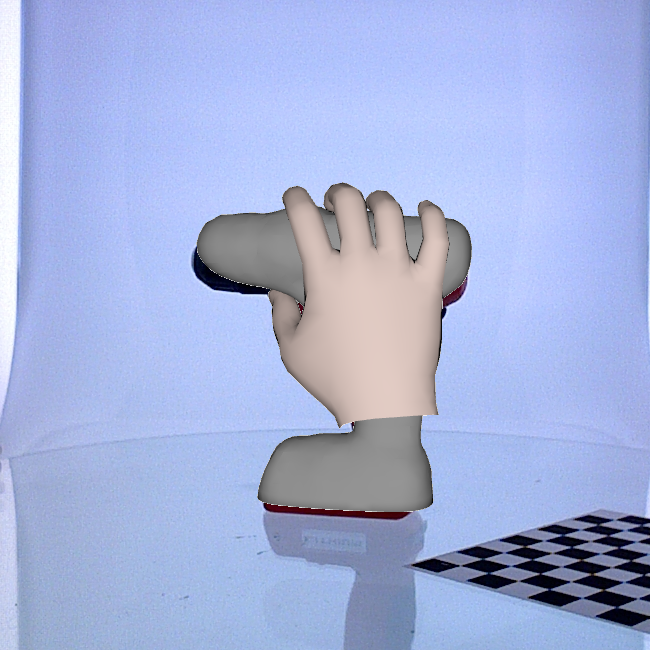}
  \end{subfigure}
  \begin{subfigure}[t]{0.15\linewidth}
    \includegraphics[width=\linewidth]{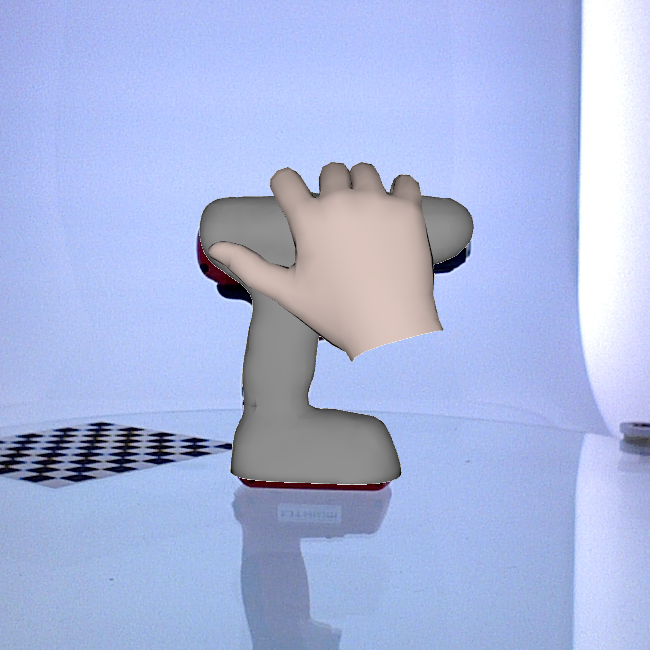}
  \end{subfigure}
  \vspace{-1.0mm}
  \caption{
  \textbf{Grasp synthesis from RGB-D.}
  We use RGB-D captures from the YCB dataset~\cite{calli2017yale} (top row) to reconstruct object models from which we synthesize grasps (bottom row).
  Our method can synthesize plausible grasps not just from ground truth object models, but also from imperfect reconstructions.
  }
  \label{fig:4-rgbd}
  \vspace{-6.0mm}
\end{figure}

\vspace{-2.0mm}
\subsection{Grasp synthesis from RGB-D input of unknown objects} \label{exp:E2}

\noindent \textbf{Setting.}
One possible application of our method is to direct grasp prediction from RGB-D images
by simulation on reconstructed object models.
Currently, our method is too slow to be used online (about 5 minutes per grasp),
but as simulation speeds increase and recent works in implicit fields push reconstruction accuracy higher and higher,
we believe that grasp prediction by simulation models will become increasingly viable.
To validate the plausibility of using our method with reconstructed object models, we present results from running our system on meshes reconstructed from RGB-D inputs.
We synthesize grasps based on RGB-D (with camera pose) inputs from the YCB object dataset~\cite{calli2017yale}.
In addition to reconstructed meshes, the YCB dataset provides the original RGB-D captures the meshes are based on.
Each object was captured from 5 different cameras at 120 different angles for a total of 600 images.
To confirm that our method can work with reconstructions done under more realistic assumptions, we limit our reconstructions to using 5 different angles from 3 cameras (2.5\% of captures).

\vspace{2.0mm}
\noindent \textbf{Data.}
For a subset of the YCB objects, we generate Poisson surface reconstructions and use our method to synthesize MANO hand grasps.
Since the inputs are from cameras with a known pose, the object reconstruction is in the world frame.
Details in the Appendix~A.4.

\vspace{2.0mm}
\noindent \textbf{Results.}
Our results confirm the viability of using simulation to synthesize grasps on reconstructed object models.
Qualitative results are presented in Figure~\ref{fig:4-rgbd}; 
additional results can be found in Appendix~D.
Although synthesis does not perform as well as with ground-truth models,
plausible human grasps are discovered for many objects
and the grasps appear well-aligned with the real-world object poses.
Future work could take advantage of learning-based reconstruction methods
to achieve grasp synthesis with fewer input images.

\begin{table}[t]
  \centering
  \resizebox{0.85\linewidth}{!} 
  {
  \begin{tabular}{l|S[table-format=4.5]S[table-format=4.5]S[table-format=4.5]S[table-format=4.5]S[table-format=4.5]S[table-format=4.5]}
  \toprule
  \rowcolor[HTML]{CBCEFB} 
  \textbf{Method} & \textbf{CA} $\uparrow$ & \textbf{IV} $\downarrow$& $\frac{\mathrm{\mathbf{CA}}}{\mathrm{\mathbf{IV}}}$ $\uparrow$ & \textbf{$\epsilon$ $\uparrow$} & \textbf{Vol $\uparrow$} & \textbf{SD $\downarrow$} \\ 
  \midrule
  \rowcolor{Lightapricot}
  Scale/Unit  & \footnotesize{$\text{cm}^2$} & \footnotesize{$\text{cm}^3$} & \footnotesize{$\text{cm}^{-1}$} & \footnotesize{$\times10^{-1}$} & \footnotesize{$\times10^{1}$} & \footnotesize{$\text{cm}$} \\ 
  \midrule
  \algoName & 42.6 & 2.83 & 15.1 & 2.38 & 20.6 & 0.41 \\
  \rowcolor[HTML]{EFEFEF} 
  \algoName w/o coarse-to-fine & 43.2 & 2.84 & 15.2 & 2.37 & 20.7 & 0.55 \\
  \algoName w/o problem relaxation & 6.1 & 0.40 & 15.2 & 0.52 & 4.0 & 3.82 \\
  \bottomrule
  \end{tabular}
  }
  \small
  \vspace{2.0mm}
  \caption{
  \textbf{Ablation study.}
  We validate our design choices with an ablation study.
  Our relaxed problem formulation has a large positive impact on all metrics.
  The quantitative impact of coarse-to-fine smoothing is more limited, but we observe a qualitative difference in grasps generated with and without smoothing.
  }
  \label{exp:results-e3}
  \vspace{-8.0mm}
\end{table}

\vspace{-1.5mm}
\subsection{Ablation study} \label{exp:E3}

We investigate the impact of our coarse-to-fine smoothing 
(Section~\ref{subsec:obj_model}), leaky contact force gradients (Section~\ref{subsec:contact-dynamics}),
and relaxed problem formulation (Section~\ref{subsec:relaxed-objective}).
We generate MANO hand grasps on 21 objects from the YCB dataset~\cite{calli2017yale}.
\emph{\algoName w/o coarse-to-fine} does not pad or smooth the object.
\emph{\algoName w/o problem relaxation} attempts to solve the problem without introducing additional force variables or a relaxed objective.
This amounts to the ``basic procedure'' described in Section~\ref{sec:grasp-metric},
i.e., directly optimize over hand pose to minimize $\Lgrasp$ and the heuristic losses.

\vspace{2.0mm}
\noindent \textbf{Results.} 
We adopt the same data as in Section~\ref{exp:E1}.
Table~\ref{exp:results-e3} presents the results.
Our relaxed problem formulation is key to our method's success, and without it, performance greatly degrades by all measures,
with discovered grasps creating very little contact (low contact area and intersection volume).
Coarse-to-fine smoothing has a modest impact, with all metrics comparable with or without smoothing,
except for simulation displacement, which is about 25\% higher without smoothing.
We did not include a variant without leaky gradient,
since this variant would never make contact with the object (if the hand is not touching the object at initialization, there will be no gradient to follow and optimization will immediately be stuck in a plateau).
\vspace{-1.0mm}
\section{Conclusions}
\vspace{-1.0mm}

We presented a simulation-based grasp synthesis pipeline capable of generating 
large datasets of plausible, high-contact grasps.
By being differentiable, our simulator is amenable to gradient-based optimization, 
allowing us to produce high-quality grasps, even for multi-finger grippers, while scaling to 
thousands of dense contacts. 
Our experiments have shown that we outperform the existing classical grasping algorithm both quantitatively and qualitatively. 
Our approach is compatible with PyTorch and can be easily integrated into existing pipelines. 
More importantly, the produced grasps can directly benefit any vision pipeline that learns grasp prediction from synthetic data.

\title{
Grasp'D: Differentiable Contact-rich Grasp Synthesis for Multi-fingered Hands \\
Supplementary Material
}

\titlerunning{Grasp'D: Differentiable Contact-rich Grasp Synthesis} 
%
\author{Dylan Turpin\inst{1,2,3} \and
Liquan Wang\inst{1,2} \and
Eric Heiden\inst{3} \and
Yun-Chun Chen\inst{1,2} \and
Miles Macklin\inst{3} \and
Stavros Tsogkas\inst{4} \and
Sven Dickinson\inst{1,2,4} \and
Animesh Garg\inst{1,2,3}
}
\authorrunning{D. Turpin et al.}
%
\institute{$^1\,$University of Toronto, $^2\,$Vector Institute, $^3\,$Nvidia, $^4\,$Samsung  \\
\email{dylanturpin@cs.toronto.edu} 
}
\maketitle

\appendix

\section*{Overview}

In this supplementary material, we provide additional details and results to complement the main paper.
Specifically:

\begin{itemize}
  \item We describe the details of our implementation and experimental setting. (Appendix~\ref{app:experiment-details}).
  \item We provide additional results of our method applied to the YCB dataset~\cite{calli2017yale} with both a human MANO hand model~\cite{MANO:SIGGRAPHASIA:2017} and a robotic Allegro hand model. (Appendix~\ref{app:ycb-results}).
  \item We provide visualizations of optimization trajectories for MANO hand grasps of YCB and ShapeNet objects, which show how grasps improve as optimization progresses. (Appendix~\ref{app:traj})
  \item We provide additional results for the validation of grasp synthesis with RGB-D reconstruction presented (Section 4.3 of the main paper). (Appendix~\ref{app:rgbd-results}).
\end{itemize}

\section{Details of implementation and experiments} \label{app:experiment-details}

\subsection{Dataset listings}

\begin{itemize}
  \item Table~\ref{tab:shapenet-objects} - ShapeNet object listing (for main experiment in section 4.2).
  \item Table~\ref{tab:ycb-objects} - YCB object listing (for rgb-d experiment in section 4.3).
  \item Table~\ref{tab:ycb-rgbd} - YCB object listing (for ablation experiment in section 4.4).
\end{itemize}

\begin{table}[t]
  \centering
  \begin{minipage}[c]{0.49\textwidth}
   \resizebox{\linewidth}{!}
  {
  \begin{tabular}{llll}
    \toprule
    \rowcolor[HTML]{CBCEFB} 
    \textbf{Category ID} & \textbf{Shape ID}  \\ 
    \midrule
    2876657 &1071fa4cddb2da2fc8724d5673a063a6 & & \\
2876657 &109d55a137c042f5760315ac3bf2c13e & & \\
2876657 &10dff3c43200a7a7119862dbccbaa609 & & \\
2876657 &10f709cecfbb8d59c2536abb1e8e5eab & & \\
2876657 &114509277e76e413c8724d5673a063a6 & & \\
2876657 &1349b2169a97a0ff54e1b6f41fdd78a & & \\
2876657 &134c723696216addedee8d59893c8633 & & \\
2880940 &12ddb18397a816c8948bef6886fb4ac & & \\
2880940 &13e879cb517784a63a4b07a265cff347 & & \\
2880940 &154ab09c67b9d04fb4971a63df4b1d36 & & \\
2880940 &18529eba21e4be8b5cc4957a8e7226be & & \\
2880940 &188281000adddc9977981b941eb4f5d1 & & \\
2880940 &1a0a2715462499fbf9029695a3277412 & & \\
2880940 &1b4d7803a3298f8477bdcb8816a3fac9 & & \\
2942699 &1298634053ad50d36d07c55cf995503e & & \\
2942699 &147183af1ba4e97b8a94168388287ad5 & & \\
2942699 &15e72ce7a8a328d1fd9cfa6c7f5305bc & & \\
2942699 &17a010f0ade4d1fd83a3e53900c6cbba & & \\
2942699 &1967344f80da29618d342172201b8d8c & & \\
2942699 &1ab3abb5c090d9b68e940c4e64a94e1e & & \\
2942699 &1cc93f96ad5e16a85d3f270c1c35f1c7 & & \\
2946921 &100c5aee62f1c9b9f54f8416555967 & & \\
2946921 &10c9a321485711a88051229d056d81db & & \\
2946921 &11c785813efc4b8630eaaf40a8a562c1 & & \\
2946921 &129880fda38f3f2ba1ab68e159bfb347 & & \\
2946921 &147901ede668deb7d8d848cc867b0bc8 & & \\
2946921 &17ef524ca4e382dd9d2ad28276314523 & & \\
2946921 &19fa6044dd31aa8e9487fa707cec1558 & & \\
2992529 &1101db09207b39c244f01fc4278d10c1 & & \\
2992529 &1105c21040f11b4aec5c418afd946fad & & \\
2992529 &112cdf6f3466e35fa36266c295c27a25 & & \\
2992529 &113303df7880cd71226bc3b9ce9ff2a1 & & \\
2992529 &11e925e3ea180b583388c2584b2f0f90 & & \\
2992529 &11f7613cae7d973fd7e59c29eb25f02f & & \\
2992529 &128bb46234d7250721844676433a0aca & & \\
3593526 &10af6bdfd126209faaf0ad030fc37d94 & & \\
3593526 &1168c9e9db2c1c5066639e628d6519b6 & & \\
3593526 &117843347cde5b502b18a5129db1b7d0 & & \\
3593526 &1252b0fc818969ebca2ed12df13a916a & & \\
3593526 &12d643221a3edaa4ab361b6be63163da & & \\
3593526 &12ec19e85b31e274725f67267e31c89 & & \\
3593526 &133dc38c1316d9515dc3653f8341633a & & \\
3624134 &102982a2159226c2cc34b900bb2492e & & \\
3624134 &118141f7d22bc46eaeb7b7328341827a & & \\
3624134 &11c987c9a34457e48c2fa4fb6bd3e62 & & \\
3624134 &135f75a374a1e22c46cb8dd27ae7fcd & & \\
3624134 &13bf5728b1f3b6cfadd1691b2083e9e7 & & \\
3624134 &13d183a44f143ca8c842482418ab083d & & \\
3624134 &1460eded8006b10139c78a1e40e247f3 & & \\
4074963 &1941c37c6db30e481ef53acb6e05e27a & & \\
4074963 &1aa78ce410bdbcd92530f02db7e9157e & & \\
4074963 &2053bdd83749adcc1e5c09d9fe5c0c76 & & \\
4074963 &226078581cd4efd755c5278938766a05 & & \\
4074963 &240456647fbca47396d8609ec76a915b & & \\
4074963 &25182f6e03375c9e7b6fd5468f603b31 & & \\
4074963 &259539bd48513e3410d32c800df6e3dd & & \\
    \bottomrule
  \end{tabular}
  }
  \end{minipage}
  \begin{minipage}[c]{0.49\textwidth}
  \vspace{-3.0mm}
  \caption{
    For experiment 1 (comparison to Obman) in Section 4.2 of the main paper, we use the following ShapeNet objects.
  }
  \label{tab:shapenet-objects}
  \end{minipage}
\end{table}

\begin{table}[t]
  \centering
  \begin{minipage}[c]{0.4\textwidth}
   \resizebox{\linewidth}{!}
  {
  \begin{tabular}{llll}
    \toprule
    \rowcolor[HTML]{CBCEFB} 
    \textbf{Object ID} & \textbf{Object Name} \\ 
    \midrule
    001 &chips\_can \\
    \rowcolor[HTML]{EFEFEF}
    002 &master\_chef\_can \\
    005 &tomato\_soup\_can \\
    \rowcolor[HTML]{EFEFEF}
    006 &mustard\_bottle \\
    008 &pudding\_box \\
    \rowcolor[HTML]{EFEFEF}
    010 &potted\_meat\_can \\
    021 &bleach\_cleanser \\
    \rowcolor[HTML]{EFEFEF}
    035 &power\_drill \\
    \bottomrule
  \end{tabular}
  }
  \end{minipage}
  \begin{minipage}[c]{0.49\textwidth}
  \vspace{-3.0mm}
  \caption{
    For experiment 2 (RGBD reconstruction) we use the following YCB objects.
  }
  \label{tab:ycb-objects}
  \end{minipage}
\end{table}

\begin{table}[t]
  \centering
  \begin{minipage}[c]{0.4\textwidth}
   \resizebox{\linewidth}{!}
  {
  \begin{tabular}{llll}
    \toprule
    \rowcolor[HTML]{CBCEFB} 
    \textbf{Object ID}  & \textbf{Object Name}  \\ 
    \midrule
    002 &master\_chef\_can \\
    \rowcolor[HTML]{EFEFEF}
003 &cracker\_box \\
004 &sugar\_box \\
\rowcolor[HTML]{EFEFEF}
005 &tomato\_soup\_can \\
006 &mustard\_bottle \\
\rowcolor[HTML]{EFEFEF}
007 &tuna\_fish\_can \\
008 &pudding\_box \\
\rowcolor[HTML]{EFEFEF}
009 &gelatin\_box \\
010 &potted\_meat\_can \\
\rowcolor[HTML]{EFEFEF}
011 &banana \\
019 &pitcher\_base \\
\rowcolor[HTML]{EFEFEF}
021 &bleach\_cleanser \\
024 &bowl \\
\rowcolor[HTML]{EFEFEF}
025 &mug \\
035 &power\_drill \\
\rowcolor[HTML]{EFEFEF}
036 &wood\_block \\
037 &scissors \\
\rowcolor[HTML]{EFEFEF}
040 &large\_marker \\
051 &large\_clamp \\
\rowcolor[HTML]{EFEFEF}
052 &extra\_large\_clamp \\
061 &foam\_brick \\
    \bottomrule
  \end{tabular}
  }
  \end{minipage}
  \begin{minipage}[c]{0.49\textwidth}
  \vspace{-3.0mm}
  \caption{
    For experiment 2 (RGBD reconstruction) we use the following YCB objects.
  }
  \label{tab:ycb-rgbd}
  \end{minipage}
\end{table}

\subsection{Initialization and smoothing schedule}
\label{app:init-and-smoothing}
\subsubsection{Initialization.}\ 
Since our grasp synthesis pipeline relies on gradient-based optimization,
the final result depends on how the parameters are initialized,
i.e., different initial hand poses will recover different final grasps.
This is a useful quality in that it allows us to sample a variety of grasps for each object by sampling different starting poses.
The force variables $\fd$ are always initialized to zero.
We employ a simple heuristic (adapted from~\cite{brahmbhatt2019contactgrasp}) to initialize the hand pose $\qh^{(0)}$.
We set all hand joints to their fully open position.
To find an initial rotation and position for the hand base link,
we uniformly sample an approach point $a$ on the object surface
and a roll angle $\theta$ around the approach vector.
We use an approach vector opposing the object surface normal at the approach point,
and set the hand rotation such that the palm's normal is aligned with the approach vector.
Finally we apply the sampled roll $\theta$ around the approach vector.
We set the hand position so that the palm's center is at a $10$cm distance from the approach point along the approach vector.
%
%
%
%

\subsubsection{Coarse-to-fine smoothing schedule.}\ 
We set the initial value for the coarse-to-fine smoothing radius $r$ to the distance between the object and the closest point on the hand  less $1$cm.
The radius is then decreased to $0$ on a linear schedule over the first 5,000 steps of a 7,000 step optimization and remains at $0$ for the final 2,000 steps.
The early steps of the optimization find a rough pose for the hand  (where on the object to grasp and an approximate finger configuration)
and the later steps optimize over fine-grained geometry, allowing the discovery of grasps that conform closely to detailed surface geometry.

\subsection{Mesh processing}

We use a discretized SDF representation as described in Section~3.2 of the main paper.
Computing the SDF involves some preprocessing.
For the experiments in Section~4.2 and 4.4 (on complete ShapeNet and YCB meshes respectively),
the input is a mesh from the relevant dataset.
For the RGB-D reconstruction experiment in Section~4.3,
the input is a reconstructed mesh (see Appendix~\ref{app:reconstruction} for details of the reconstruction pipeline).
To compute the sign of the SDF at a given query point, we must determine whether that point is inside or outside the object.
This is more straightforward if the mesh consists of a single closed surface, so we first run ManifoldPlus~\cite{huang2020manifoldplus} to compute a watertight mesh.
Next, a ($256 \times 256 \times 256$) grid of points is evenly sampled over the mesh bounding box (padded by $1$cm)
and the signed distance of each point to the mesh is computed using libigl~\cite{jacobson2017libigl}.

\subsection{Reconstruction pipeline}
\label{app:reconstruction}
We describe the RGB-D reconstruction pipeline used in Section 4.3.
The YCB object dataset includes RGB-D captures for each object.
The object is placed on a spinning platter surrounded by 5 cameras
and is captured at each of 120 different angles as the plate is rotated in 3 degree increments.
We take 15 of these depth images (captures from the first, third and fifth camera at 5 angles in $72$ degree increments).
We run the code provided alongside the YCB dataset
in order to register the depth maps and combine them into a single
world frame point cloud.
We create a Poisson reconstruction~\cite{kazhdan2006poisson} of this point cloud
using the Open3D library~\cite{Zhou2018} with a depth of $5$.
The resulting mesh is still incomplete because the bottom of the object is not visible (since it is the contact surface between the table and the object)
We use PyMeshFix~\cite{sullivan2019pyvista} to complete this and any other remaining holes in the mesh.

\subsection{Simulation details}
We run each simulation for a single timestep of length $1\times10^{-5}$ seconds.
For the MANO hand model, all $773$ vertices are used as contact locations.
For the Allegro hand model, we sample $\sim$3000 points on the surface to use as contact locations.
In all experiments we set the normal stiffness to $\kn = 1\times10^{6}$, frictional stiffness to $\kf = 1\times10^{8}$, and the friction coefficient to $\mu = 0.8$.
For the leaky gradient (described in Section~3.3 of the main paper) we set the proportion of gradient that leaks through non-colliding contact forces to $\alpha=0.1$.
Note that the above applies to simulation during grasp optimization.
When we compute simulation displacement for evaluation purposes,
we do not use our own simulator,
but instead use PyBullet~\cite{coumans2021} (details in Appendix~\ref{app:eval}).

\subsection{Evaluation details}
\label{app:eval}
We evaluate grasps in terms of their contact patterns and stability.

\subsubsection{Interpenetration volume}\ is the volume of the intersection between the hand and the object.
Lower values are better (since in reality the hand cannot penetrate the hard object).
We compute this by voxelizing the hand (with 1mm resolution) and querying the object's SDF at each voxel position to decide if each hand voxel is overlapping the object or not.

\subsubsection{Contact area}\ is the area of surface contact (in $cm^2$) between the hand and the object.
This is computed similarly to interpenetration volume, except that only the hand surface is voxelized 
(i.e., the hand is treated as an empty shell, not a solid volume).

\subsubsection{Contact area to interpenetration volume ratio.}\ 
Interpenetration can be avoided by simply avoiding contact with the object entirely,
so there is a trade off between interpenetration volume and the other metrics.
To capture the amount of interpenetration, conditional on the amount of contact, we report the ratio of contact area to interpenetration.

\subsubsection{$\epsilon$ (Ferrari-Canny) metric}\ 
measures grasp stability using the magnitude of the smallest force that can break a grasp.
A more stable grasp can withstand larger forces, so a larger force will be needed to break the grasp.
This quantity is equivalent to the size of the largest origin-centered ball contained in the Grasp Wrench Space (GWS~\cite{ferrari1992planning}).
The GWS is the space of wrenches the contacts induced by the grasp can withstand,
assuming that the total hand-object wrench will be a linear combination of the wrenches at each contact with coefficients summing to $1$.
Under a Coulomb friction model, the possible wrenches at each contact are defined by a friction cone, which we approximate by a pyramid.
\subsubsection{Volume metric}\  is an alternate measure of stability that considers all the possible forces a grasp can withstand (instead of just the magnitude of the smallest force that breaks the grasp).
The volume metric is simply the volume of the GWS.

\subsubsection{Simulation displacement}\ is a simulation-based, rather than analytic, measure of stability.
We use GANHand's implementation~\cite{corona2020ganhand} of a simulation displacement metric in PyBullet~\cite{coumans2021} to measure grasp stability by checking how far the object is displaced from its initial pose when the grasp is applied.
We use the default physics parameters provided by GANHand except for setting the friction coefficient to $1.2$ (instead of the seemingly high default of $3.0$).
Whereas we train (i.e., optimize) our grasps in our own custom simulator, this metric is computed in a widely used third-party simulator (PyBullet), with a different collision detector, contact model and time stepping scheme.
This avoids giving ourselves an unfair advantage by training (optimizing) and testing (computing evaluation metrics)
with the same contact model and physics engine (which baselines we compare to did not have access to).

\subsubsection{True signed distances.}\ 
Whenever a metric relies on the object SDF (e.g., to compute contact forces or to determine if a voxel is intersecting the object or not), we compute that SDF with libigl~\cite{jacobson2017libigl} using the ground truth mesh instead of a discrete grid approximation of the SDF.

\subsubsection{Evaluating on top $k$ grasps.}\ 
For our method, we report metrics for the top 2 and top 5 grasps (ordered by simulation displacement -- details below) for each object.
For the ObMan dataset, we report the top 2 and top 5 grasps for each object (ordered by their heuristic measure, described below).
The ObMan generation procedure uses the GraspIt! simulator to synthesize grasps by optimizing (with simulated annealing) over an analytic metric.
%
%
Many grasps for each object are generated by running about 70k steps annealing steps.
The top 2 grasps are then selected according to a heuristic measure (see Appendix C.2 of~\cite{hasson2019learning})
which encourages palm and phalange contact.
This heuristic was explicitly added to compensate for the bias of analytic synthesis towards fingertip-only grasps.
To test our own method, we generate 10 grasps for each object, each using 7000 optimization steps.
We report these metrics over the top 2 and top 5 grasps with the lowest simulation displacement.

\subsection{Optimization details}
We used the ADAMax~\cite{zhang2018improved} optimizer to update the the hand pose parameters $\qh^{(0)}$ (with a learning rate of  $3\times10^{-3}$) and force parameters $\fd$ (with a learning rate of $1\times10^{-2}$).
Some objectives are more important than others, so are treated as constraints to satisfy rather than costs to minimize.
Specifically, we use the Modified Differential Multiplier Method~\cite{platt1987constrained}, treating $\Ltask < C_\textrm{task}$ and $\Lqlimit < C_\textrm{limit}$ as constraints,
while minimizing $\Lphysics$, $\Lqrange$ and $\Linter$.
We set $C_\textrm{task}=1\times10^{-4}$ and $C_\textrm{limit} = 1\times10^{-4}$.
Damping is set to $1.0$ for all constraints.
During MANO hand experiments,
we do not use the joint limit loss $\Lqlimit$ or joint limit constraint, as these limits appear to be well-handled implicitly by the PCA parameterization.
Similarly, we do not compute the self-intersection loss for the MANO hand, yet recover grasps with low self-intersection due to the hand parameterization.
All losses are used enabled for the Allegro hand.

\subsection{Timing}
On a mobile Nvidia RTX 2070, generating a MANO hand grasp for a YCB object (by taking 7,000 optimizer steps) takes about 5 minutes.
The MANO hand has only 773 vertices, so the memory footprint of the simulation is limited and three grasps can be synthesized in parallel, reducing average grasp synthesis time to about 2 minutes.
While not yet approaching realtime performance, this is comparable to the speed of analytic synthesis with the GraspIt! simulator~\cite{miller2004graspit}, which takes around 5 minutes~\cite{corona2020ganhand} to synthesize a grasp when using the eigengrasp planner with simulated annealing as for the ObMan dataset~\cite{hasson2019learning}.

\section{YCB results}
\label{app:ycb-results}
\begin{figure}[!t]
    \centering
    \includegraphics[width=\linewidth]{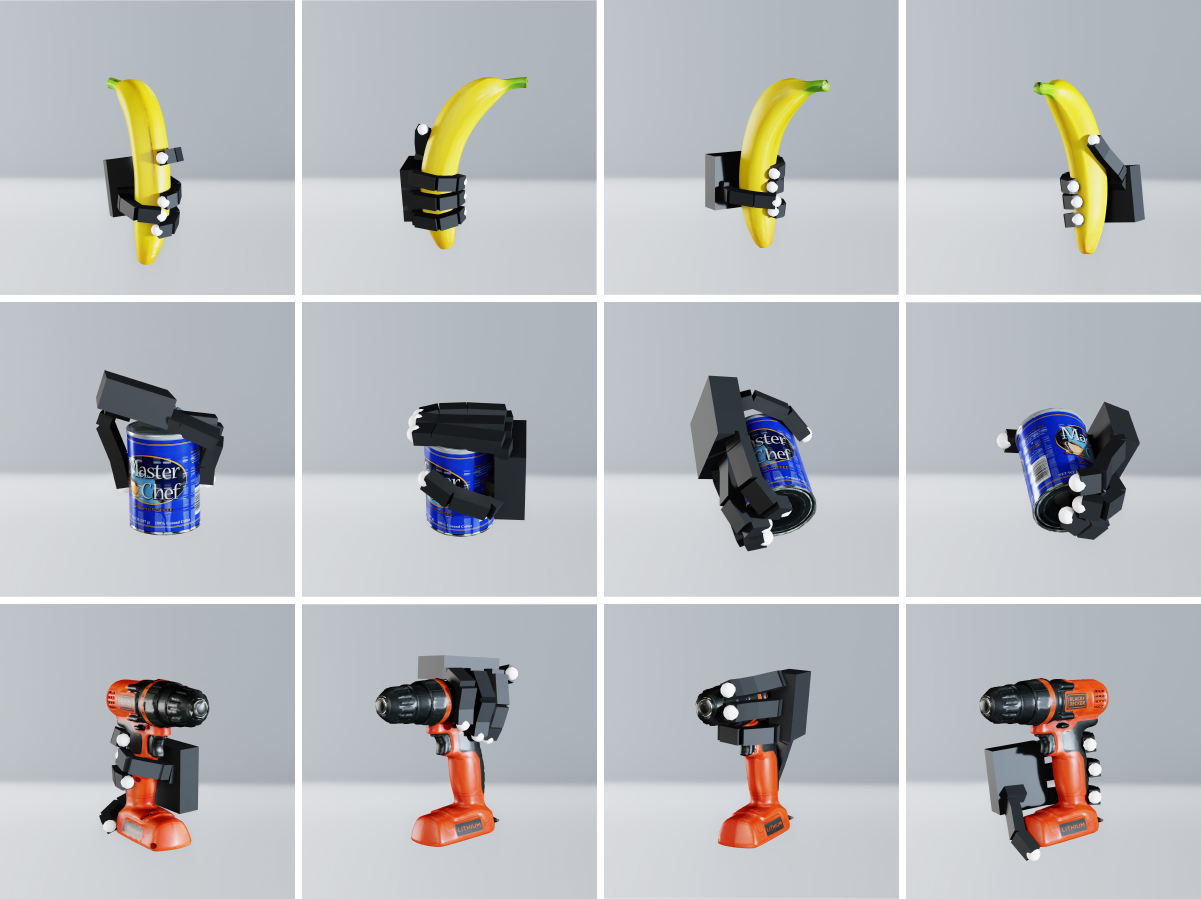}
    \caption{
    \textbf{Robotic grasping with the four-fingered Allegro hand.}
    Our method works equally well with robotic and human hand models.
    We visualise grasps of three YCB objects~\cite{calli2017yale} with the four-fingered Allegro robotic hand.
    We can recover a variety of grasps for each object by sampling different initial hand poses (which gradient-based optimization takes to different final grasps).
    See Appendix~\ref{app:init-and-smoothing} for details of initialization.
    }
    \label{fig:1-allegro}
\end{figure}

\begin{figure}[!t]
    \centering
    \includegraphics[width=\linewidth]{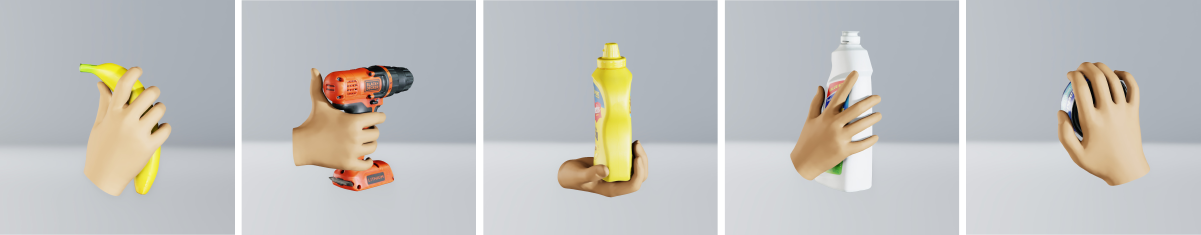}
    \caption{
    \textbf{Synthesized MANO hand grasps of YCB objects.}
    Our method generates contact-rich grasps for objects from the YCB dataset.
    These qualitative results are drawn from the ablation study in Section 4.4 of the main paper,
    specifically with all features turned on, corresponding to the row labelled Grasp'D in Table 3.
    }
    \label{fig:2-manoycb}
\end{figure}

We provide additional examples of applying our method to objects from the YCB dataset with both the Allegro robotic hand and the MANO human hand models (see Figures~\ref{fig:1-allegro} and \ref{fig:2-manoycb} respectively).

\section{Optimization trajectories}
\label{app:traj}
\begin{figure}[!t]
    \centering
    \includegraphics[width=\linewidth]{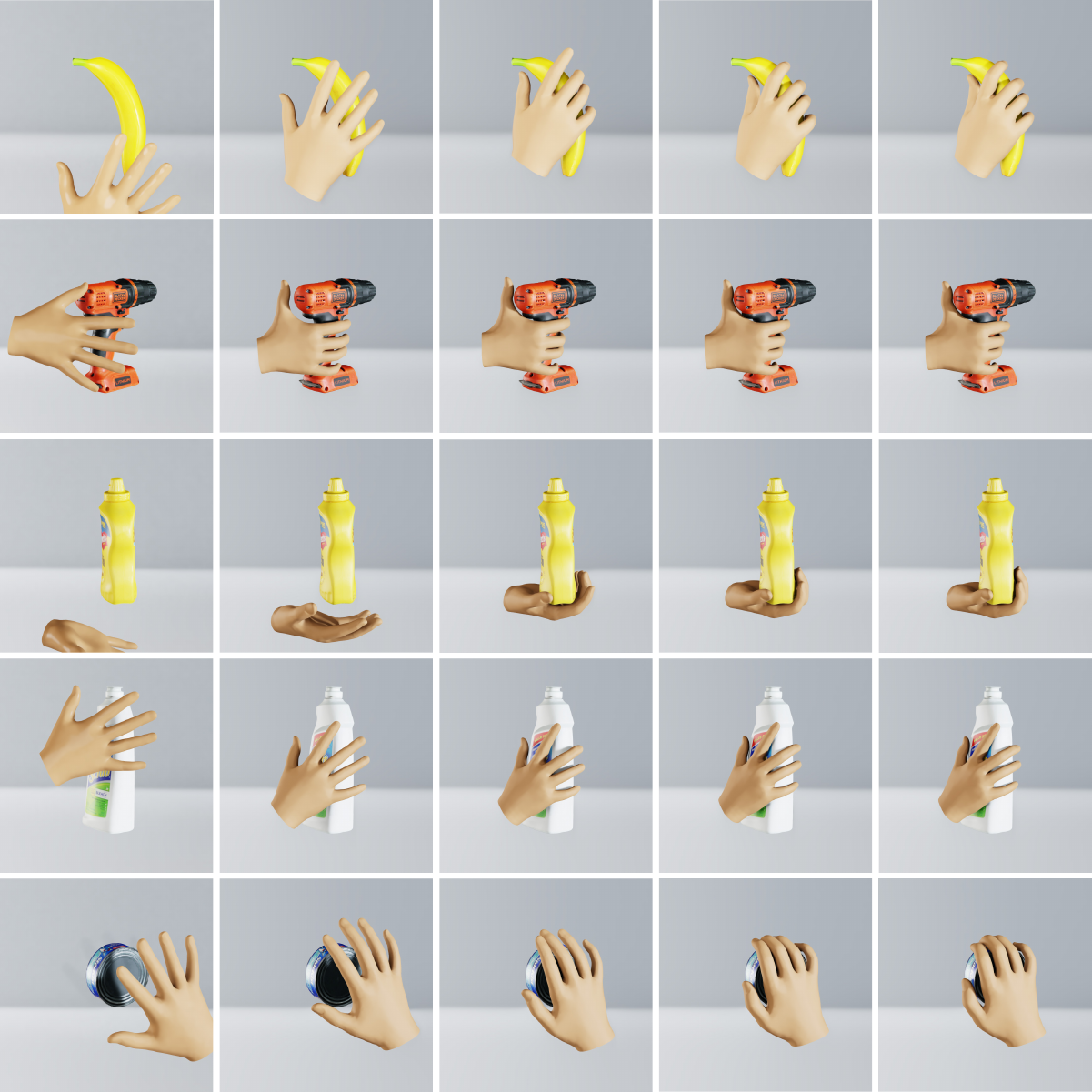}
    \caption{
    \textbf{Optimization trajectories for MANO hand grasps of YCB objects.}
    Grasps improve as optimization progresses (from left-to-right in the figure).
    We visualize the optimization paths that result in the final grasps in Figure~\ref{fig:2-manoycb}.
    The leftmost column shows the initial hand pose (see Appendix~\ref{app:init-and-smoothing} for details of initialization) and the optimization progresses from left to right until reaching the final grasps in the rightmost column.
    Initially, the hand is not even in contact with the object,
    but as optimization continues the grasp becomes higher contact, more plausible, and more stable.
    }
    \label{fig:3-manoycbtraj}
\end{figure}

\begin{figure}[!t]
    \centering
    \includegraphics[width=\linewidth]{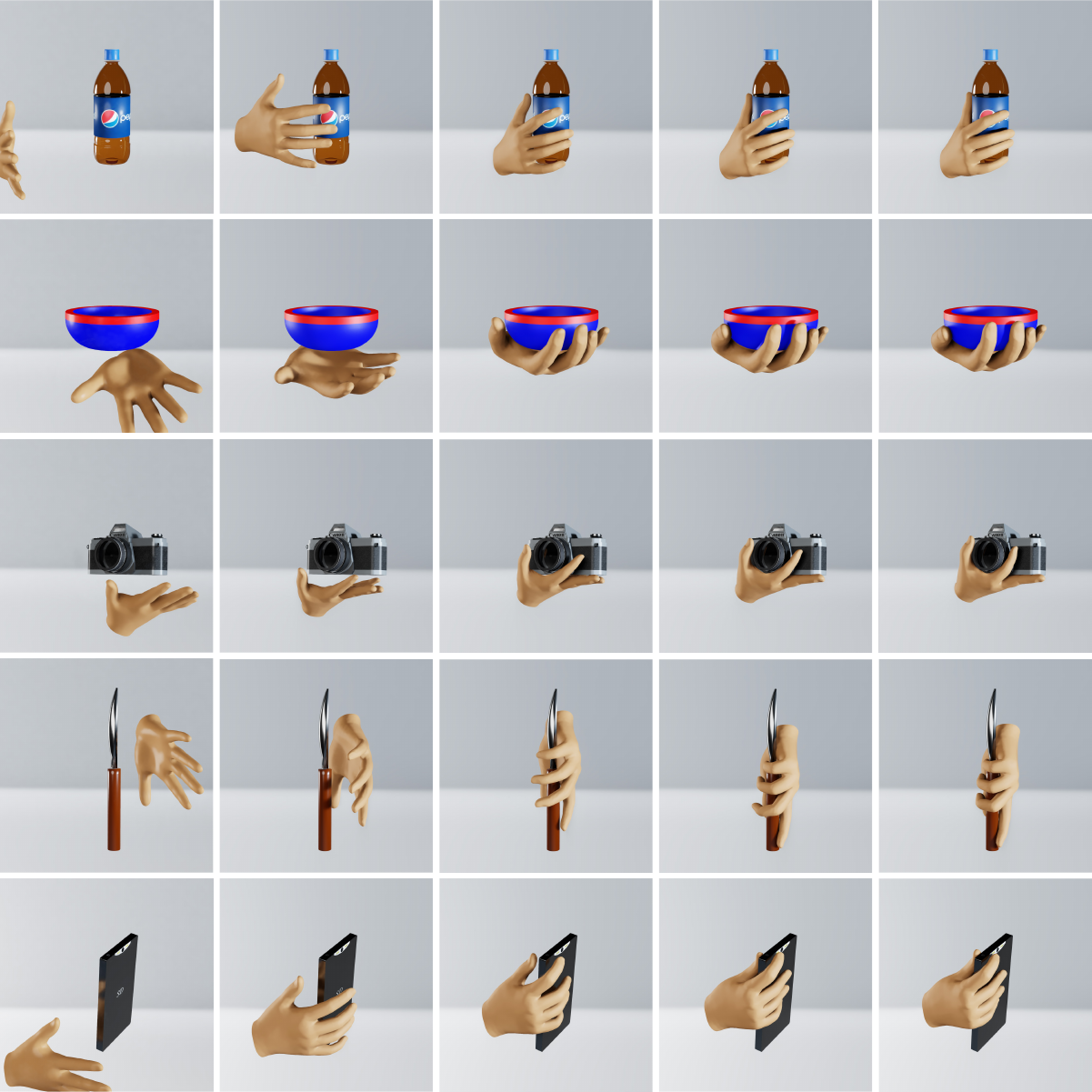}
    \caption{
    \textbf{Optimization trajectories for MANO hand grasps of ShapeNet objects.}
    Grasps improve as optimization progresses (from left-to-right in the figure).
    We visualize the optimization paths that result in the final grasps in the second row of Figure~1 of the main paper.
    }
    \label{fig:4-manoshapenettraj}
\end{figure}

To visualize how grasps evolve as optimization progresses, we render hand poses at regular intervals throughout optimization trajectories for 10 grasps (5 for objects from YCB, 5 for objects from ShapeNet -- see Figure~\ref{fig:3-manoycbtraj} and Figure~\ref{fig:4-manoshapenettraj} respectively).

\section{Additional RGB-D results}
\label{app:rgbd-results}
\begin{figure}[t]
  \centering
  \begin{subfigure}[t]{0.155\linewidth}
    \includegraphics[width=\linewidth]{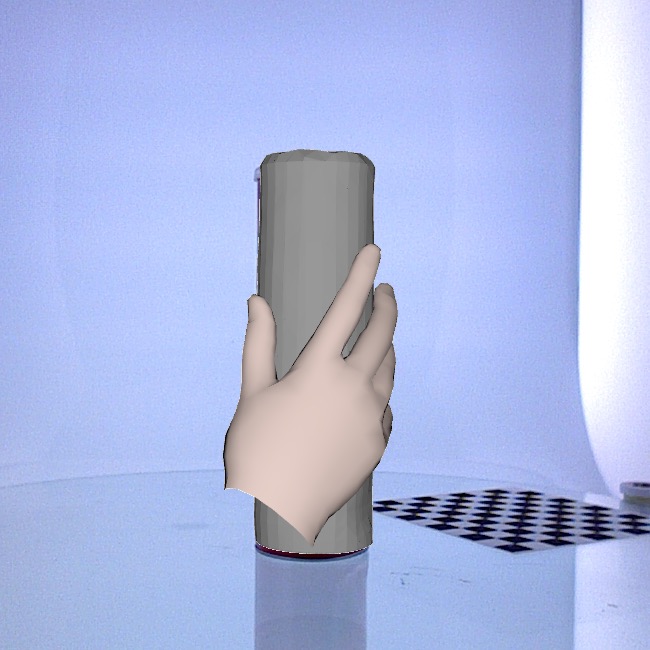}
  \end{subfigure}
  \begin{subfigure}[t]{0.155\linewidth}
    \includegraphics[width=\linewidth]{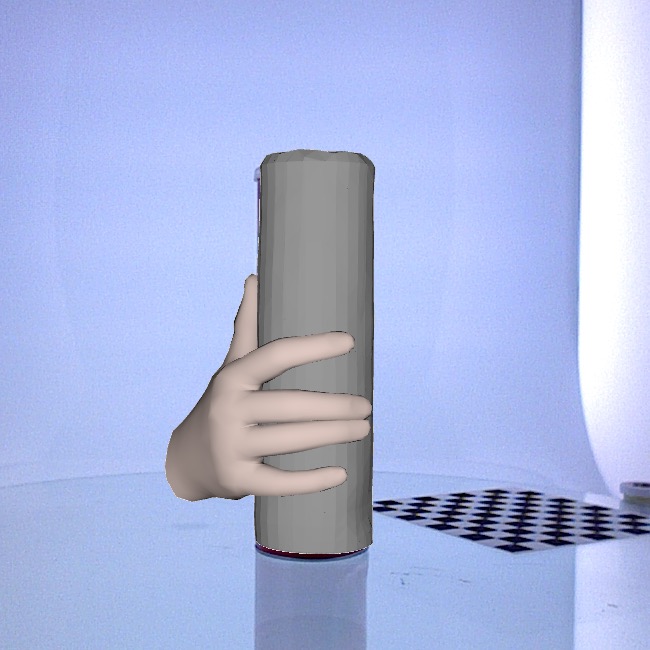}
  \end{subfigure}
  \begin{subfigure}[t]{0.155\linewidth}
    \includegraphics[width=\linewidth]{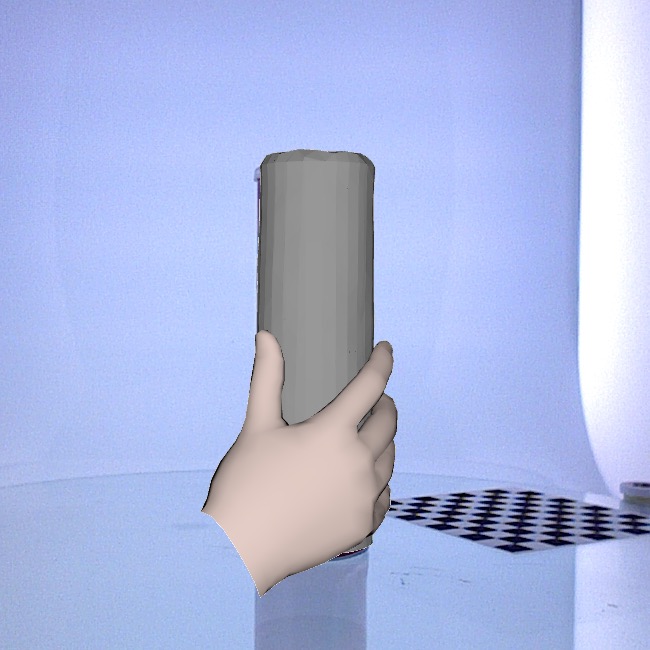}
  \end{subfigure}
  \begin{subfigure}[t]{0.155\linewidth}
    \includegraphics[width=\linewidth]{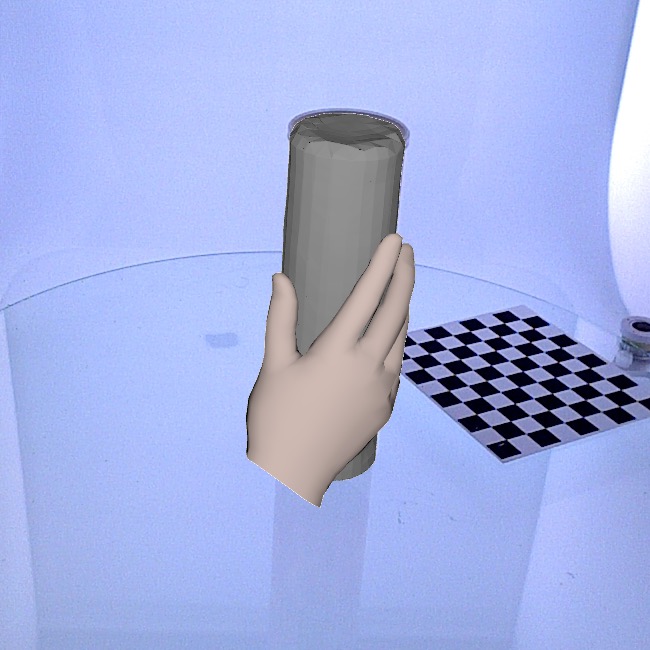}
  \end{subfigure}
  \begin{subfigure}[t]{0.155\linewidth}
    \includegraphics[width=\linewidth]{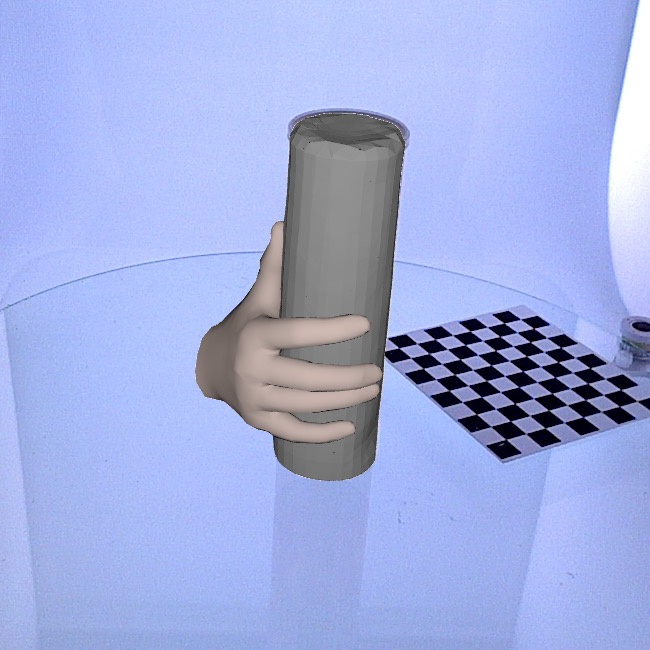}
  \end{subfigure}
  \begin{subfigure}[t]{0.155\linewidth}
    \includegraphics[width=\linewidth]{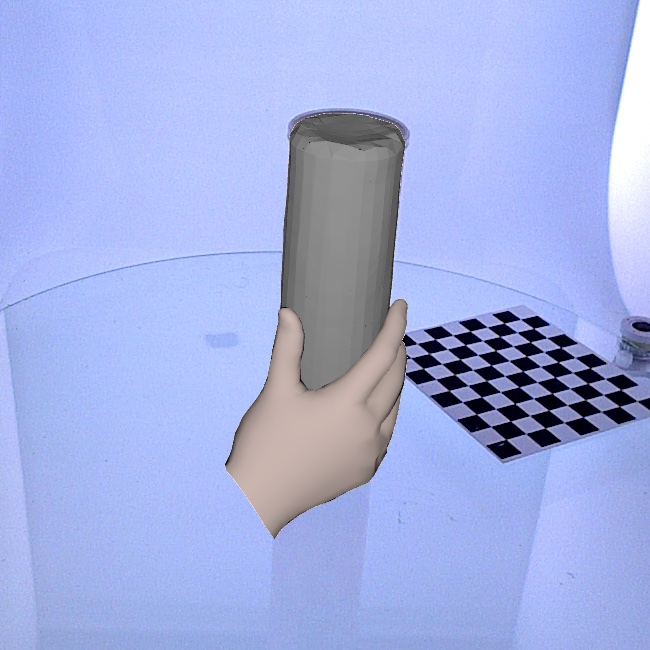}
  \end{subfigure}
  \\ 
  \vspace{1.0mm}
  \begin{subfigure}[t]{0.155\linewidth}
    \includegraphics[width=\linewidth]{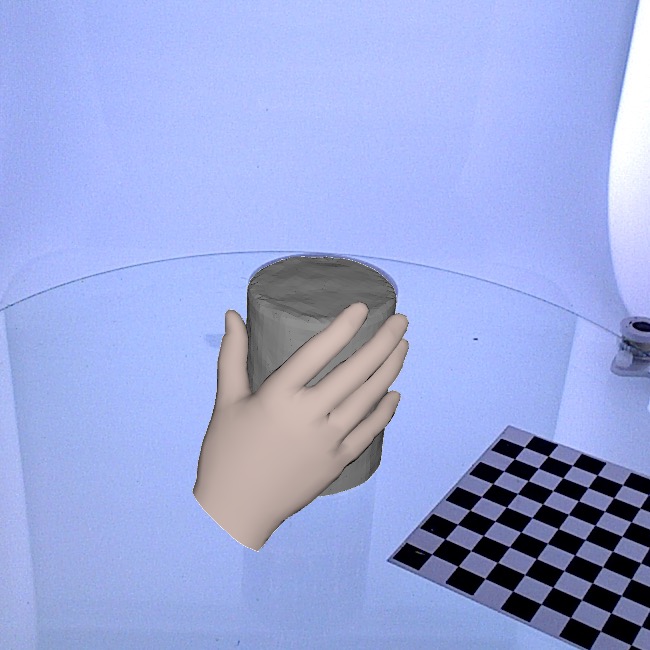}
  \end{subfigure}
  \begin{subfigure}[t]{0.155\linewidth}
    \includegraphics[width=\linewidth]{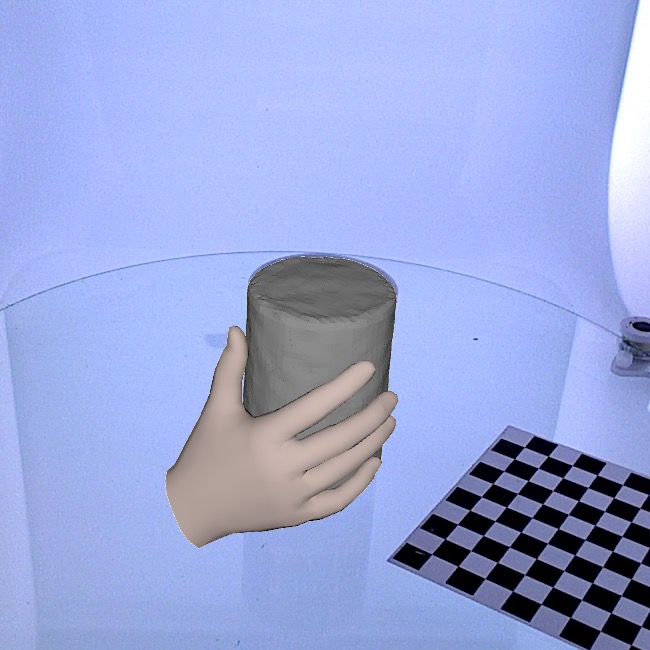}
  \end{subfigure}
  \begin{subfigure}[t]{0.155\linewidth}
    \includegraphics[width=\linewidth]{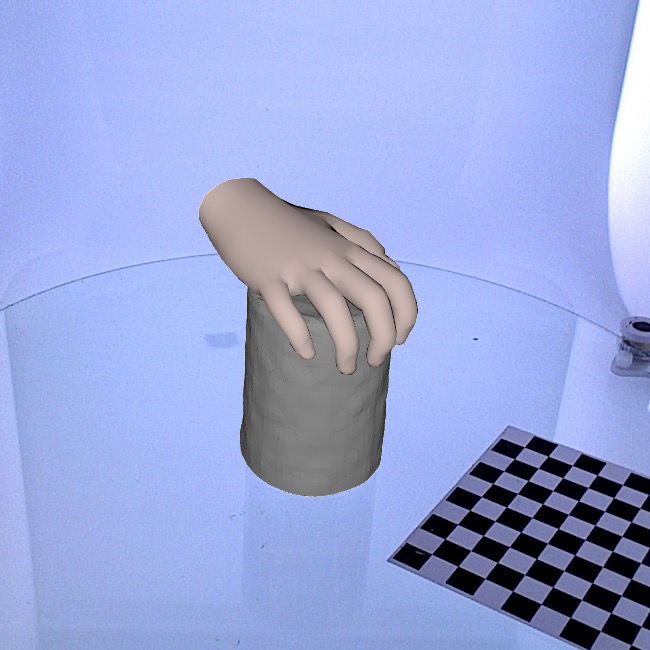}
  \end{subfigure}
  \begin{subfigure}[t]{0.155\linewidth}
    \includegraphics[width=\linewidth]{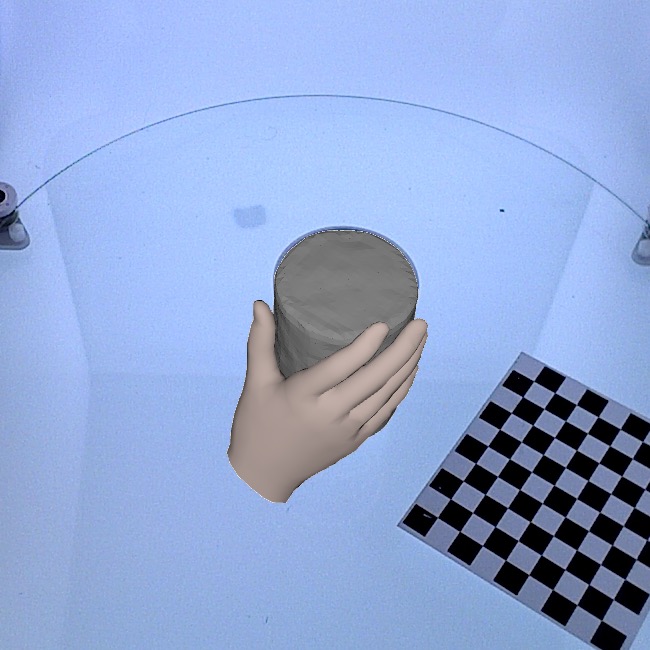}
  \end{subfigure}
  \begin{subfigure}[t]{0.155\linewidth}
    \includegraphics[width=\linewidth]{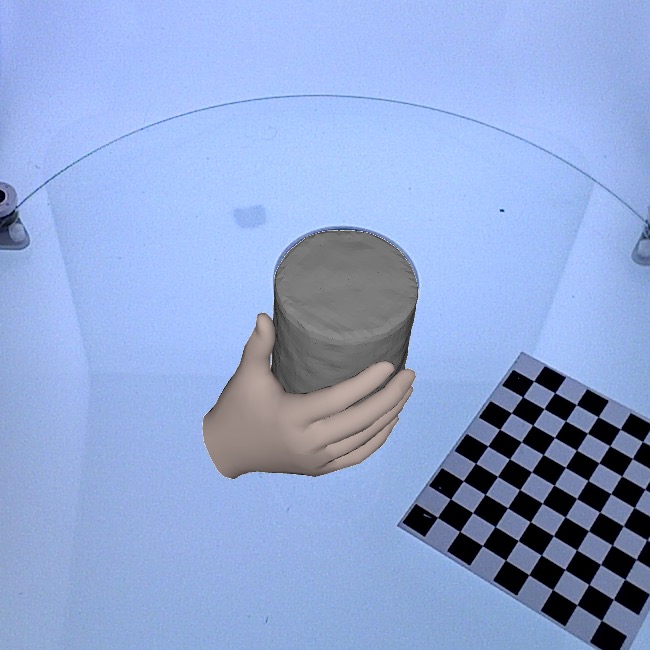}
  \end{subfigure}
  \begin{subfigure}[t]{0.155\linewidth}
    \includegraphics[width=\linewidth]{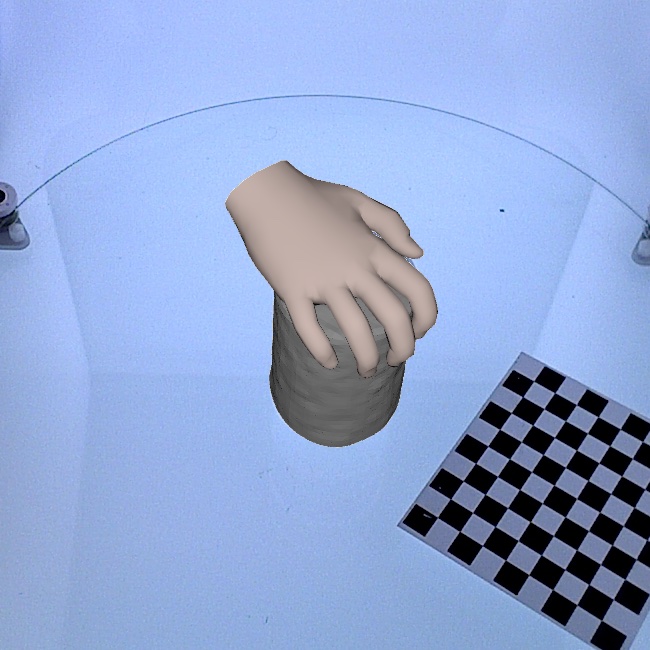}
  \end{subfigure}
  \\
  \vspace{1.0mm}
  \begin{subfigure}[t]{0.155\linewidth}
    \includegraphics[width=\linewidth]{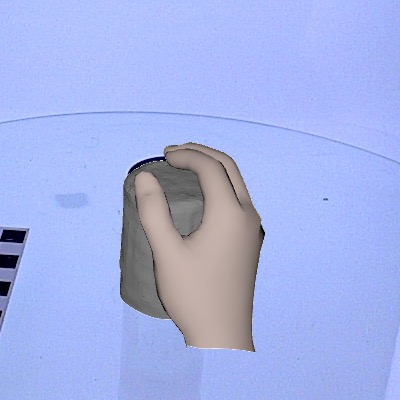}
  \end{subfigure}
  \begin{subfigure}[t]{0.155\linewidth}
    \includegraphics[width=\linewidth]{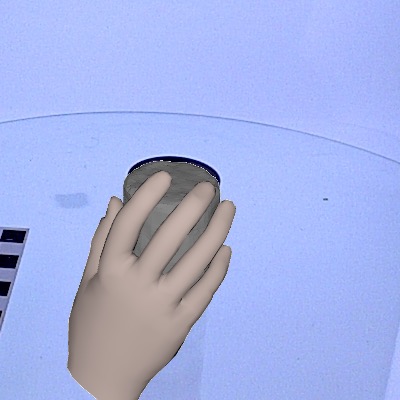}
  \end{subfigure}
  \begin{subfigure}[t]{0.155\linewidth}
    \includegraphics[width=\linewidth]{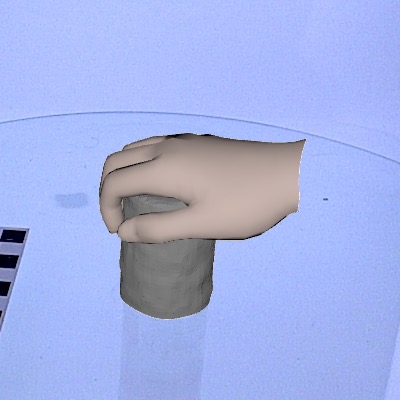}
  \end{subfigure}
  \begin{subfigure}[t]{0.155\linewidth}
    \includegraphics[width=\linewidth]{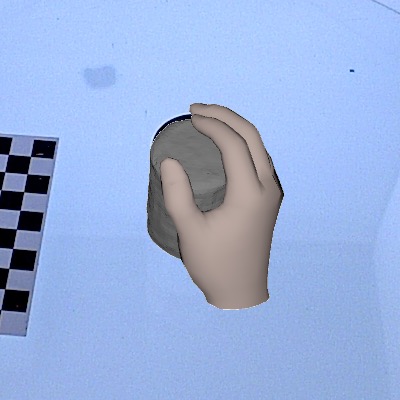}
  \end{subfigure}
  \begin{subfigure}[t]{0.155\linewidth}
    \includegraphics[width=\linewidth]{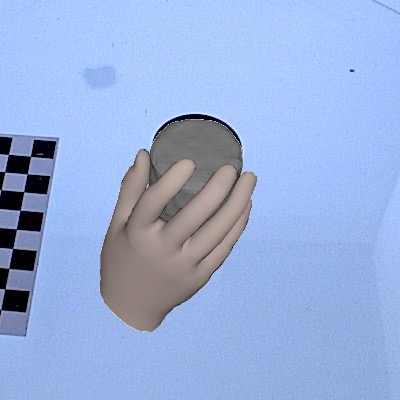}
  \end{subfigure}
  \begin{subfigure}[t]{0.155\linewidth}
    \includegraphics[width=\linewidth]{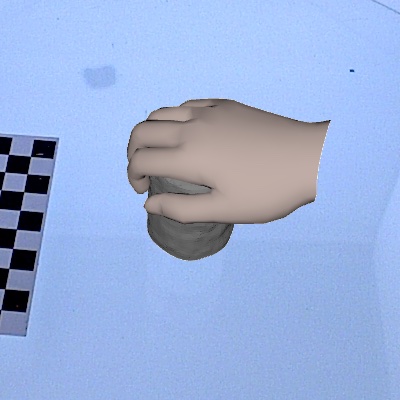}
  \end{subfigure}
  \\
  \vspace{1.0mm}
  \begin{subfigure}[t]{0.155\linewidth}
    \includegraphics[width=\linewidth]{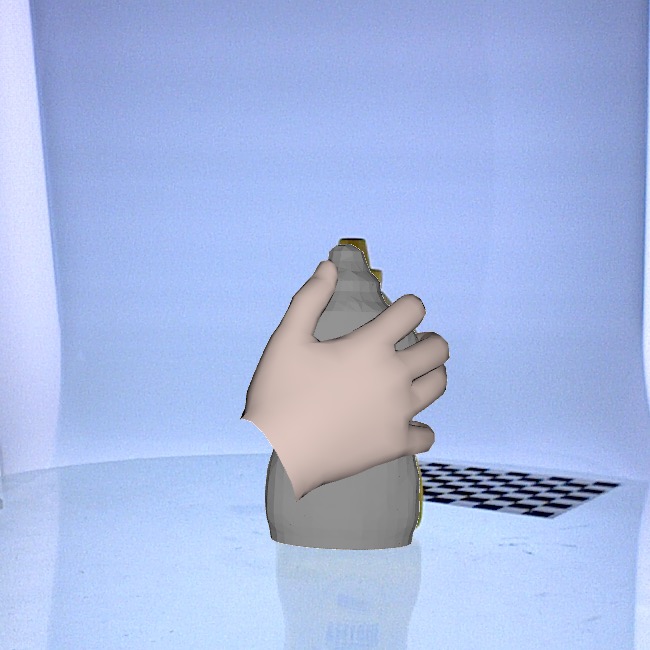}
  \end{subfigure}
  \begin{subfigure}[t]{0.155\linewidth}
    \includegraphics[width=\linewidth]{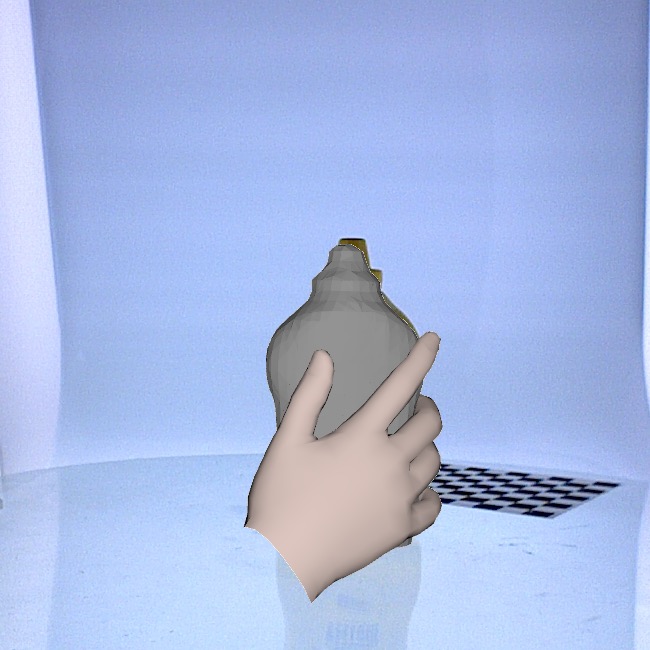}
  \end{subfigure}
  \begin{subfigure}[t]{0.155\linewidth}
    \includegraphics[width=\linewidth]{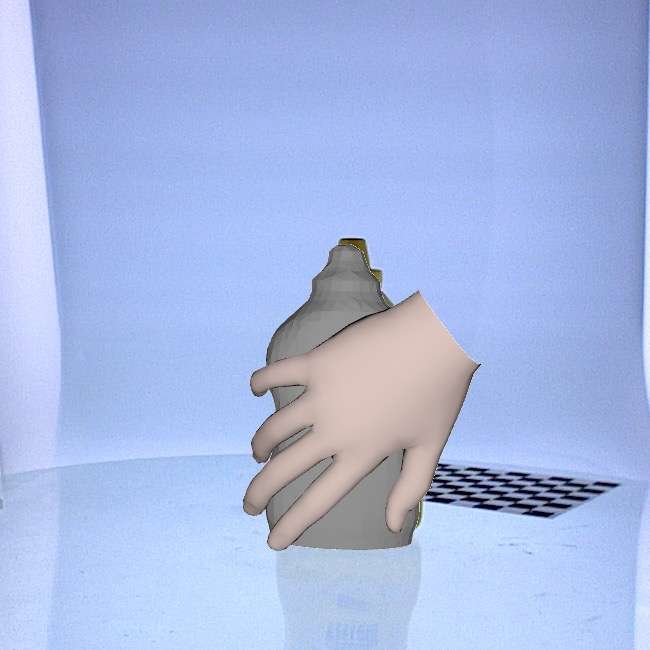}
  \end{subfigure}
  \begin{subfigure}[t]{0.155\linewidth}
    \includegraphics[width=\linewidth]{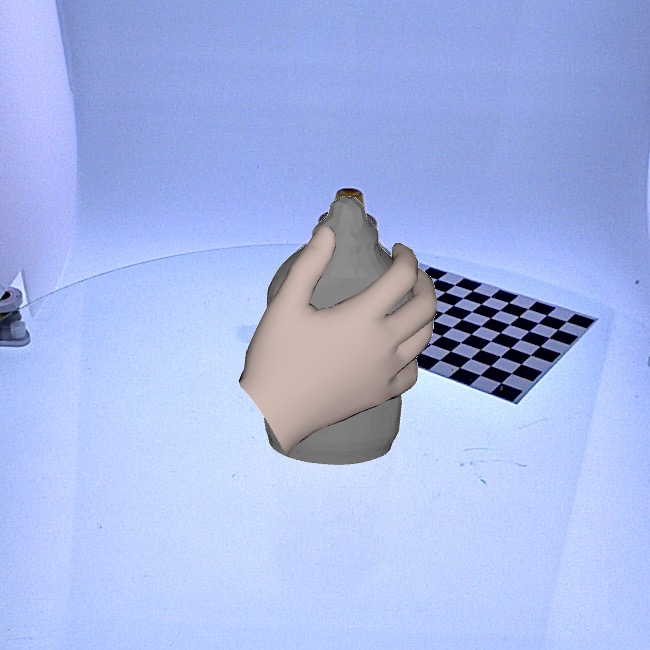}
  \end{subfigure}
  \begin{subfigure}[t]{0.155\linewidth}
    \includegraphics[width=\linewidth]{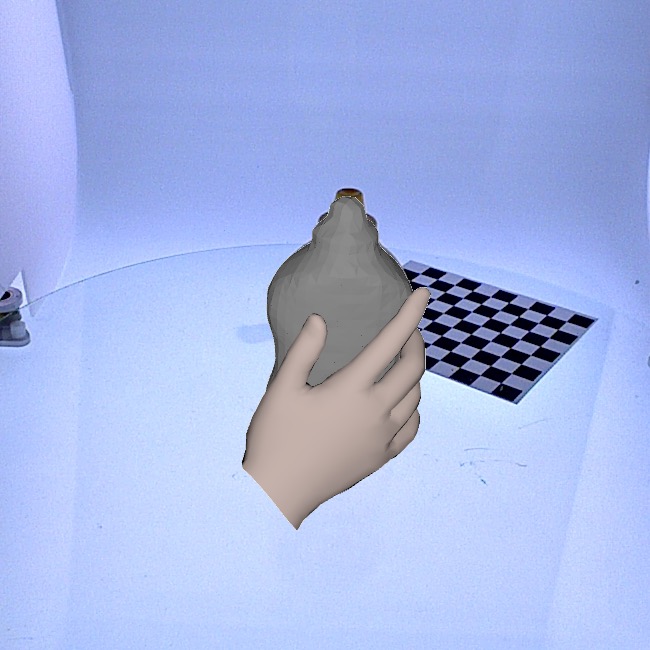}
  \end{subfigure}
  \begin{subfigure}[t]{0.155\linewidth}
    \includegraphics[width=\linewidth]{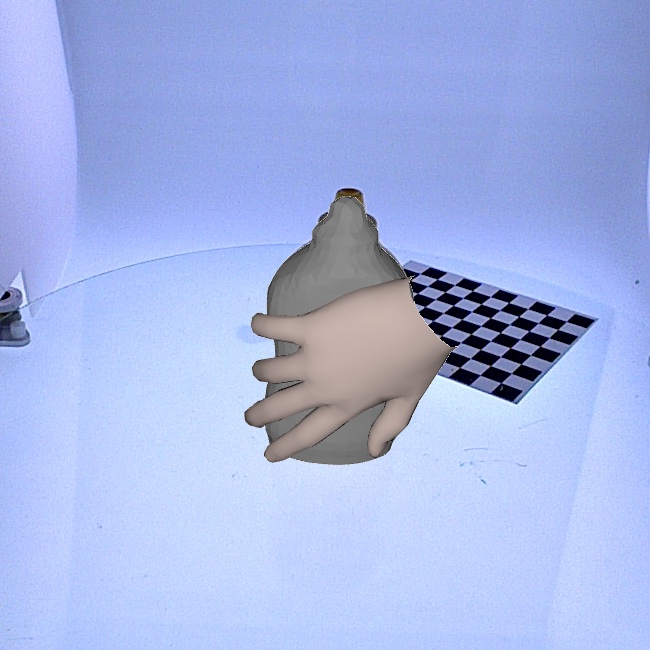}
  \end{subfigure}
  \\
  \vspace{1.0mm}
  \begin{subfigure}[t]{0.155\linewidth}
    \includegraphics[width=\linewidth]{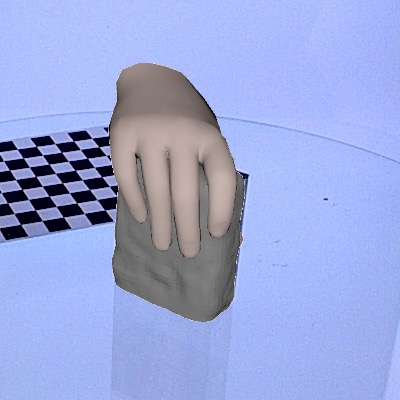}
  \end{subfigure}
  \begin{subfigure}[t]{0.155\linewidth}
    \includegraphics[width=\linewidth]{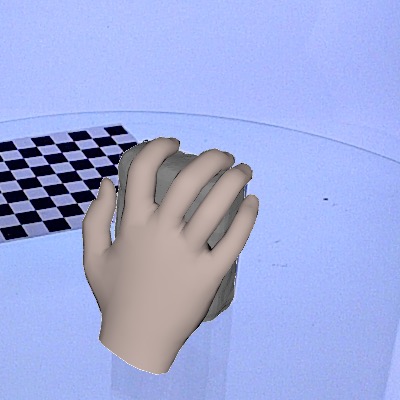}
  \end{subfigure}
  \begin{subfigure}[t]{0.155\linewidth}
    \includegraphics[width=\linewidth]{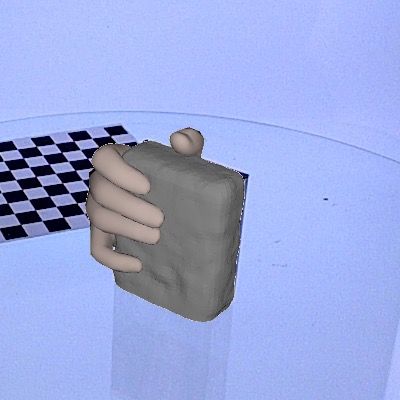}
  \end{subfigure}
  \begin{subfigure}[t]{0.155\linewidth}
    \includegraphics[width=\linewidth]{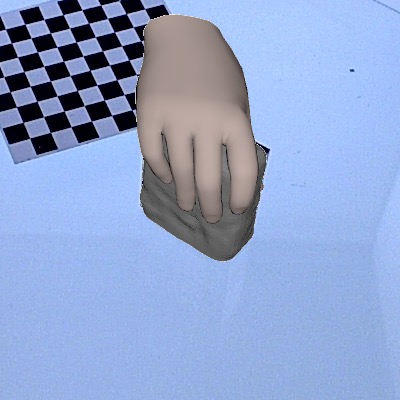}
  \end{subfigure}
  \begin{subfigure}[t]{0.155\linewidth}
    \includegraphics[width=\linewidth]{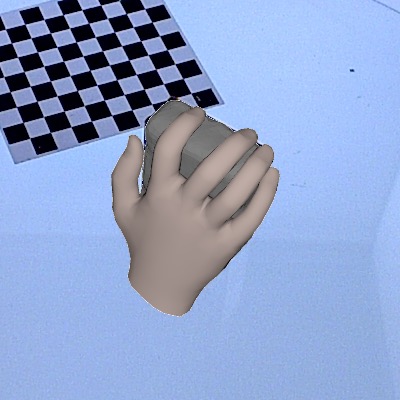}
  \end{subfigure}
  \begin{subfigure}[t]{0.155\linewidth}
    \includegraphics[width=\linewidth]{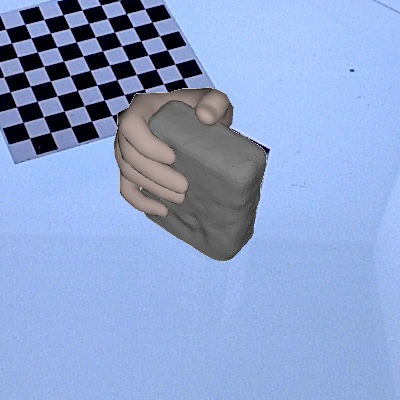}
  \end{subfigure}
  \\
  \vspace{1.0mm}
  \begin{subfigure}[t]{0.155\linewidth}
    \includegraphics[width=\linewidth]{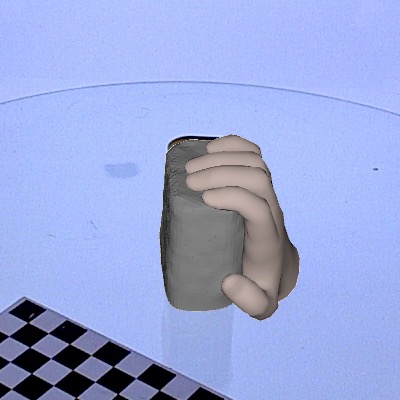}
  \end{subfigure}
  \begin{subfigure}[t]{0.155\linewidth}
    \includegraphics[width=\linewidth]{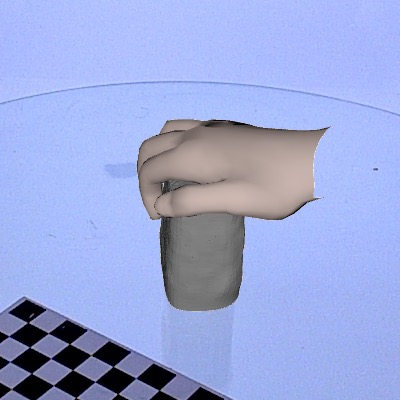}
  \end{subfigure}
  \begin{subfigure}[t]{0.155\linewidth}
    \includegraphics[width=\linewidth]{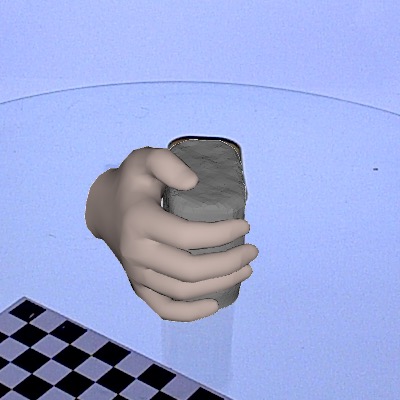}
  \end{subfigure}
  \begin{subfigure}[t]{0.155\linewidth}
    \includegraphics[width=\linewidth]{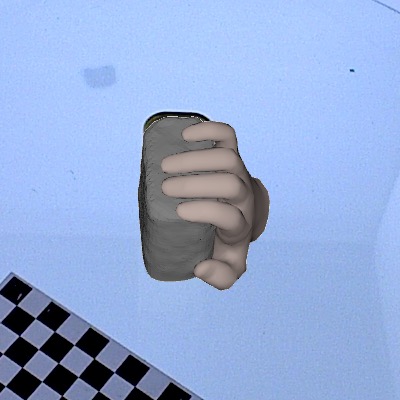}
  \end{subfigure}
  \begin{subfigure}[t]{0.155\linewidth}
    \includegraphics[width=\linewidth]{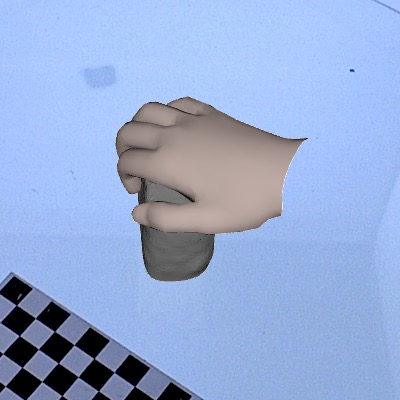}
  \end{subfigure}
  \begin{subfigure}[t]{0.155\linewidth}
    \includegraphics[width=\linewidth]{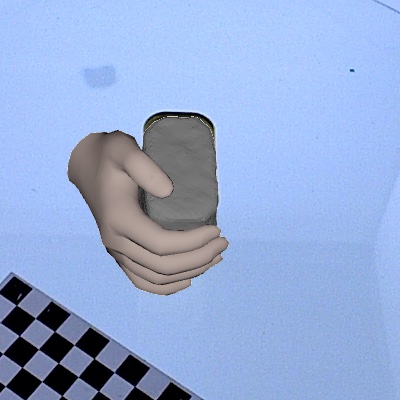}
  \end{subfigure}
  \\
  \vspace{1.0mm}
  \begin{subfigure}[t]{0.155\linewidth}
    \includegraphics[width=\linewidth]{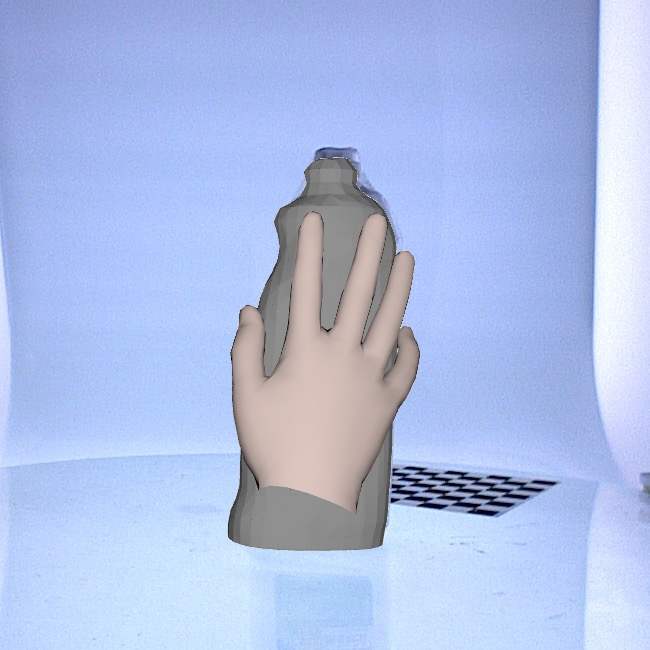}
  \end{subfigure}
  \begin{subfigure}[t]{0.155\linewidth}
    \includegraphics[width=\linewidth]{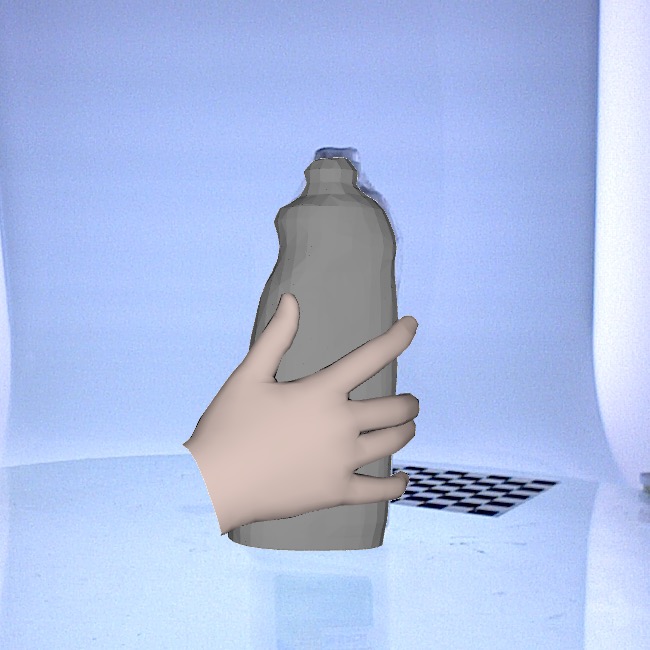}
  \end{subfigure}
  \begin{subfigure}[t]{0.155\linewidth}
    \includegraphics[width=\linewidth]{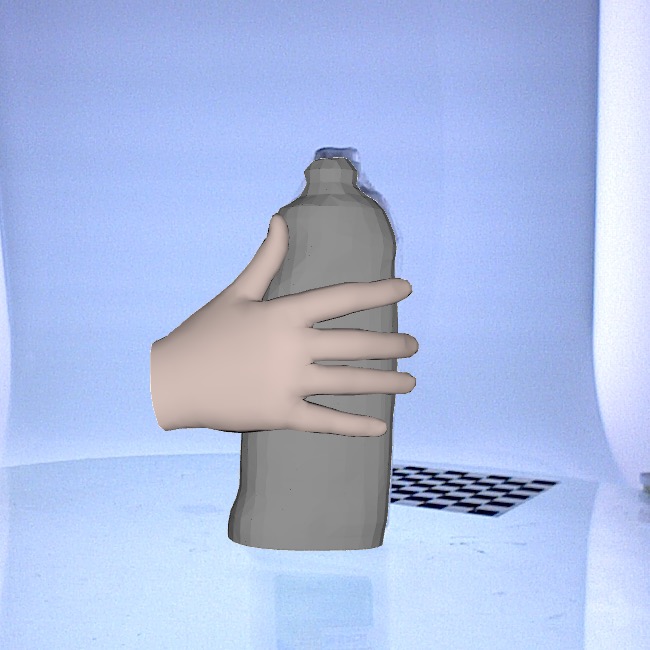}
  \end{subfigure}
  \begin{subfigure}[t]{0.155\linewidth}
    \includegraphics[width=\linewidth]{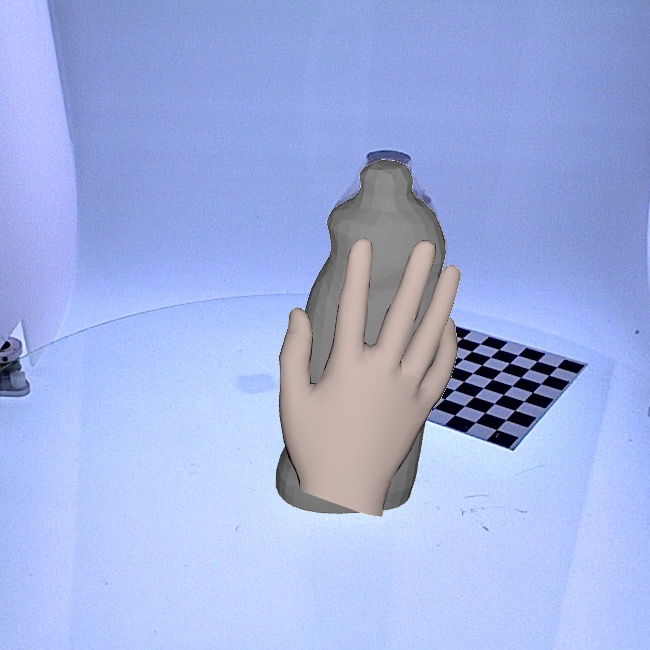}
  \end{subfigure}
  \begin{subfigure}[t]{0.155\linewidth}
    \includegraphics[width=\linewidth]{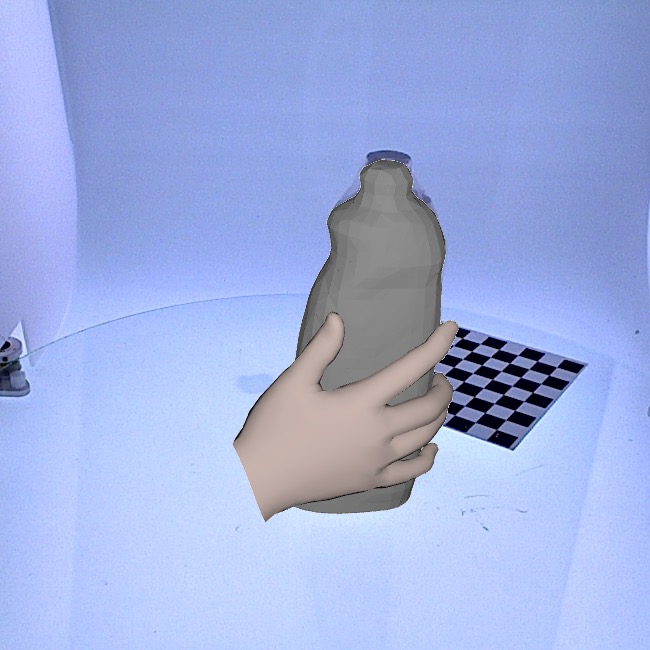}
  \end{subfigure}
  \begin{subfigure}[t]{0.155\linewidth}
    \includegraphics[width=\linewidth]{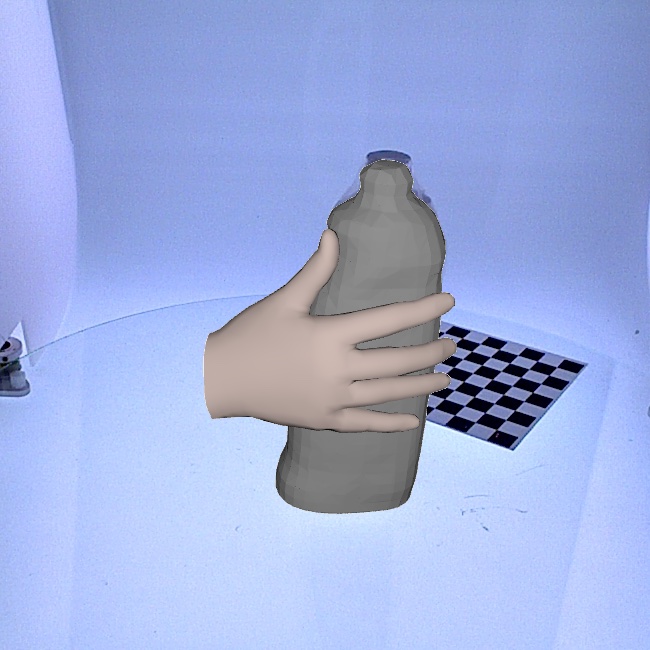}
  \end{subfigure}
  \\
  \vspace{1.0mm}
  \begin{subfigure}[t]{0.155\linewidth}
    \includegraphics[width=\linewidth]{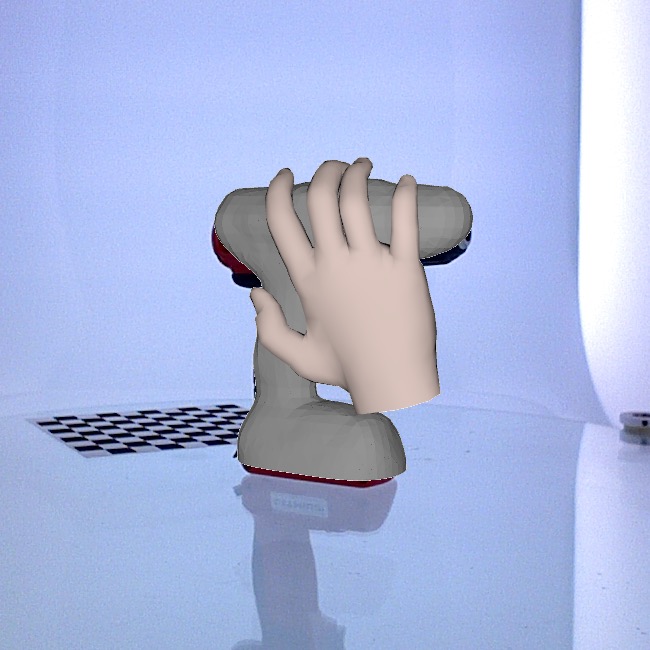}
  \end{subfigure}
  \begin{subfigure}[t]{0.155\linewidth}
    \includegraphics[width=\linewidth]{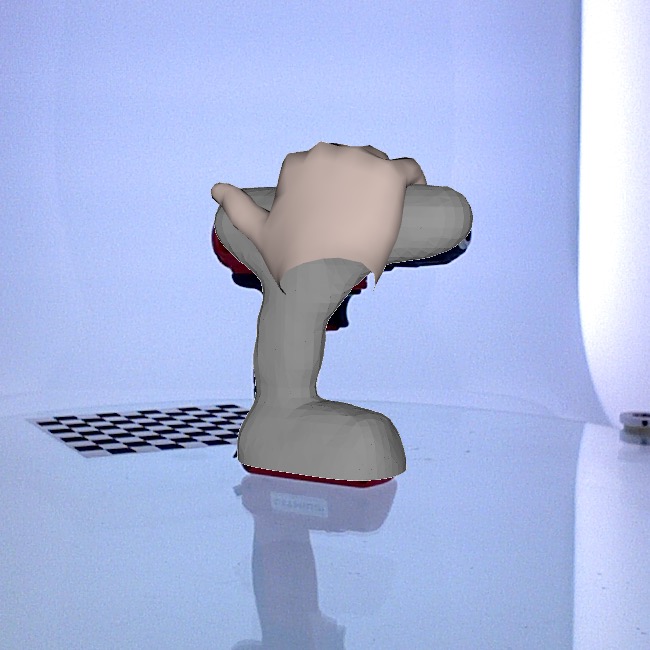}
  \end{subfigure}
  \begin{subfigure}[t]{0.155\linewidth}
    \includegraphics[width=\linewidth]{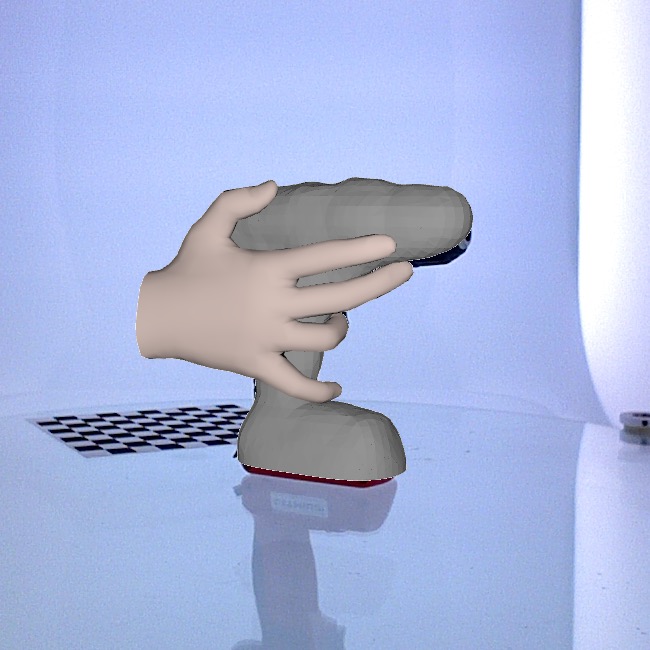}
  \end{subfigure}
  \begin{subfigure}[t]{0.155\linewidth}
    \includegraphics[width=\linewidth]{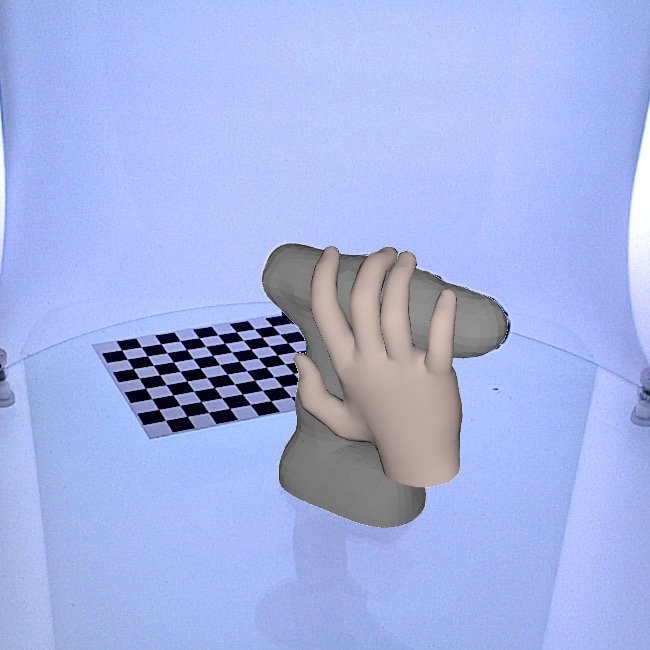}
  \end{subfigure}
  \begin{subfigure}[t]{0.155\linewidth}
    \includegraphics[width=\linewidth]{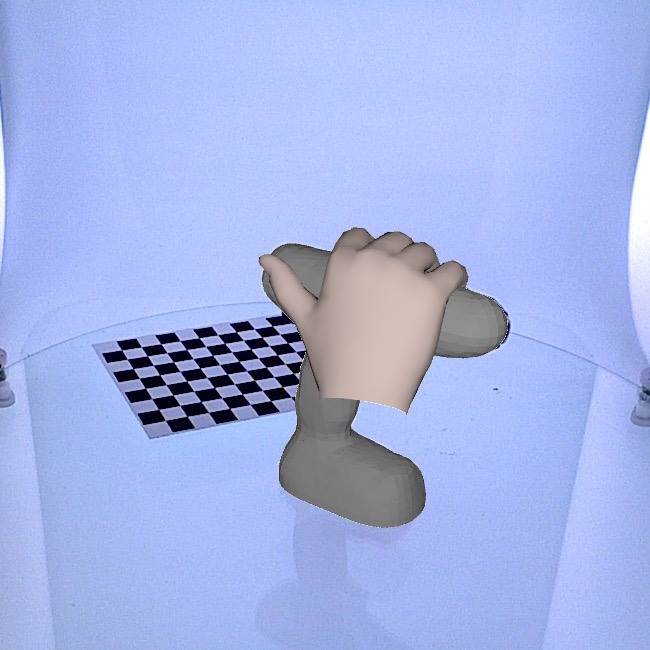}
  \end{subfigure}
  \begin{subfigure}[t]{0.155\linewidth}
    \includegraphics[width=\linewidth]{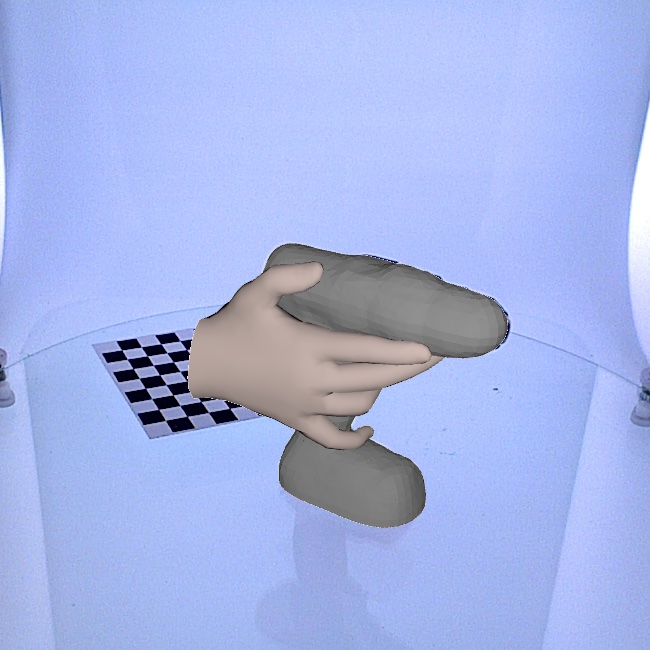}
  \end{subfigure}
  \\
  \vspace{1.0mm}
  \caption{
  \textbf{Grasp synthesis from RGB-D.}
  We use RGB-D captures from the YCB dataset~\cite{calli2017yale} to reconstruct object models from which we synthesize grasps (see section~4.3 of the main paper for details).
  Our method can synthesize plausible grasps not just from ground truth object models, but also from imperfect reconstructions.
  }
  \label{fig:ycb-rgbd}
\end{figure}

We provide additional qualitative results of applying our methods to objects reconstructed from RGBD images in the YCB dataset.
Figure~\ref{fig:ycb-rgbd} shows 3 synthesized grasps for each object visualized from 2 different viewpoints.
We also provide quantitative results for our RGB-D experiment (section~4.3 of the main paper) in Table~\ref{tab:exp1-rgbd}.

\begin{table}[H]
  \centering
  \begin{tabular}{l|S[table-format=4.5]S[table-format=4.5]S[table-format=4.5]S[table-format=4.5]}
  \toprule
  \rowcolor[HTML]{CBCEFB} 
  \textbf{Input} & \textbf{CA} $\uparrow$ & \textbf{IV} $\downarrow$ & $\frac{\mathrm{\mathbf{CA}}}{\mathrm{\mathbf{IV}}}$ $\uparrow$ &  \textbf{SD $\downarrow$} \\ 
  \midrule
  GT-Mesh & \textbf{42.6} & \textbf{2.83} & \textbf{15.1} &\textbf{0.41}\\
  \rowcolor[HTML]{EFEFEF} 
  RGB-D & 25.46 & 7.82 & 3.26 & 5.68\\
  \bottomrule
  \end{tabular}
  \vspace{1.5mm}
  \small
  \caption{
    \textbf{RGB-D experiment quantitative results.} Performance with reconstructions is poorer than with ground truth object models, but the resulting grasps are still visually plausible (see Figure~4 of the main paper, Figure~5 of the supplemental) and metrics are comparable to Grasping Field~\cite{karunratanakul2020grasping} and GANHand~\cite{corona2020ganhand}.
    Small errors in reconstruction may produce large errors in grasp synthesis, a drawback future work might address by optimizing over the reconstruction alongside the grasp.
  }
  \label{tab:exp1-rgbd}
\end{table}

\section{Training on synthesized data}
Fine-tuning Grasping Field~\cite{karunratanakul2020grasping} with data generated by Grasp'D improves performance on unseen YCB objects better than additional GraspIt!~\cite{miller2004graspit} data.
Table~\ref{tab:exp2-retrain} displays the result of fine-tuning a pre-trained Grasping Field network with additional data synthesized by either the GraspIt! simulator or Grasp'D.
%
%
The network is first trained for 1400 epochs on the ObMan dataset~\cite{hasson2019learning} (of synthetic GraspIt! grasps) and then fine-tuned for 100 epochs on 1000 new grasps (of ShapeNet~\cite{chang2015shapenet} objects already included in the 
ObMan dataset) before final testing on 8 objects from the YCB set.
Fine-tuning with Grasp'D data results in significantly higher-contact grasps.
This comes with a slight increase in intersection volume, but the ratio of contact area to intersection is improved, as is the simulation displacement.

\begin{table}[H]
  \centering
  \begin{tabular}{l|S[table-format=4.5]S[table-format=4.5]S[table-format=4.5]S[table-format=4.5]}
  \toprule
  \rowcolor[HTML]   {CBCEFB} 
  \textbf{Data source} & \textbf{CA} $\uparrow$ & \textbf{IV} $\downarrow$ & $\frac{\mathrm{\mathbf{CA}}}{\mathrm{\mathbf{IV}}}$ $\uparrow$ &  \textbf{SD $\downarrow$} \\ 
  \midrule
  GraspIt!~[58] & 15.78	 & \textbf{9.44}	 & 1.67	 & 3.30\\
  \rowcolor[HTML]{EFEFEF} 
  Grasp'D & \textbf{21.00}	& 11.23	& \textbf{1.78} & 	\textbf{2.88}\\
  \bottomrule
  \end{tabular}
  \vspace{1.5mm}
  \small
  \caption{
  \textbf{Fine-tuning Grasping Field~\cite{karunratanakul2020grasping}} with data generated by Grasp'D improves performance on unseen YCB objects better than additional GraspIt!~\cite{miller2004graspit} data.}
  \label{tab:exp2-retrain}
\end{table}

\newpage

\vspace{1.0mm}
{\small \noindent\textbf{Acknowledgements.} 
DT was supported in part by a Vector research grant. The authors appreciate the support of NSERC, Vector Institute and Samsung AI. 
AG was also supported by NSERC Discovery Grant, NSERC Exploration Grant, CIFAR AI Chair, XSeed Discovery Grant from University of Toronto. 
}
\small
\bibliographystyle{splncs04}
\bibliography{references}

\end{document}